\pdfoutput=1

\documentclass[11pt]{article}

\usepackage{emnlp2021}

\usepackage{times}
\usepackage{latexsym}
\usepackage{textcomp}

\usepackage[T1]{fontenc}

\usepackage[utf8]{inputenc}

\usepackage{microtype}

\usepackage{times}
\usepackage{latexsym}
\usepackage{multirow}
\usepackage{soul}

\usepackage{graphicx}
\usepackage{color}
\usepackage{amsmath}

\usepackage{amssymb}
\usepackage[normalem]{ulem}
\usepackage{amsthm}

\usepackage{xspace}
\usepackage{subfigure}
\usepackage{booktabs}
\usepackage{array,tabularx}
\usepackage[ruled,vlined]{algorithm2e}
\usepackage{algpseudocode}
\usepackage{amsfonts}

\usepackage{xspace}
\usepackage{microtype}
\usepackage[ruled,vlined]{algorithm2e}
\usepackage{algpseudocode}
\usepackage{microtype}

\title{Think about it! Improving defeasible reasoning by first modeling the question scenario}

\author{Aman Madaan, \textbf{Niket Tandon$^\dagger$}, Dheeraj Rajagopal,  \textbf{Peter Clark$^\dagger$},\\ 
\textbf{Yiming Yang}, \textbf{Eduard Hovy} \\
  Language Technologies Institute, Carnegie Mellon University, Pittsburgh, PA, USA \\ 
  $^\dagger$ Allen Institute for Artificial Intelligence, Seattle, WA, USA \\ 
  \texttt{\{dheeraj,amadaan,yiming,hovy\}@cs.cmu.edu} \\ \texttt{\{nikett, peterc\}@allenai.org} \\}

\date{}
\definecolor{Red}{rgb}{1,0,0}
\definecolor{Green}{rgb}{0.4,1,0.2}
\definecolor{Blue}{rgb}{0,0,1}
\definecolor{Red}{rgb}{0.9,0,0}
\definecolor{Orange}{rgb}{1,0.5,0}
\definecolor{yellow}{rgb}{0.65,0.6,0}
\definecolor{cadmiumgreen}{rgb}{0.2, 0.7, 0.24}
\newcommand{\V}[1]{\mathbf{#1}}
\newcommand{\C}[1]{\mathcal{#1}}

\newcommand{\secref}[1]{\S\ref{#1}}

\newcommand{\eat}[1]{}

\newcommand{\X}[1]{\mathbf{\mathrm{S}}}
\newcommand{\Z}[1]{\mathbf{\mathrm{C^-}}}
\newcommand{\VV}[1]{\mathbf{\mathrm{C^+}}}
\newcommand{\W}[1]{\mathbf{\mathrm{M^-}}}
\newcommand{\U}[1]{\mathbf{\mathrm{S^-}}}
\newcommand{\Y}[1]{\mathbf{\mathrm{M^+}}}
\newcommand{\LL}[1]{\mathbf{\mathrm{H^-}}}
\newcommand{\M}[1]{\mathbf{\mathrm{H^+}}}
\newcommand{\proscript}[1]{ProScript}

\newcommand{\rdim}[1]{\in \mathbb{R}^{#1}}

\newcommand{\wiqa}{\textsc{wiqa}\xspace}

\newcommand{\hq}{$\V{h}_\V{x}$\xspace}
\newcommand{\hv}{$\V{h}_\V{v}$\xspace}
\newcommand{\hg}{$\V{h}_\V{G}$\xspace}
\newcommand{\nodeset}{\V{h}_{\V{V}}}

\newcommand{\qa}{\textsc{qa}\xspace}
\newcommand{\ours}{\textsc{curious}\xspace}  

\newcommand{\lm}{$\mathcal{L}$\xspace}

\newcommand\ABox[2]{
  \fbox{\lower0.75cm
    \vbox to 1.5cm{\vfil
      \hbox to 2.1cm{\hfil\parbox{2.9cm}{#1\\#2}\hfil}
      \vfil}%
  }%
}

\newcommand{\gcn}{\textsc{gcn}\xspace}
\newcommand{\moe}{MoE\xspace}

\newcommand{\badg}{$\V{G}'$\xspace}
\newcommand{\betterg}{$\V{G}^*$\xspace}
\newcommand{\badtobetter}{$\V{G}' \rightarrow \V{G}^*$\xspace}

\newcommand{\geninit}{$\textsc{gen}_{\text{init}}$\xspace}
\newcommand{\genwitht}{$\textsc{gen}_{\text{init}}^{*}$\xspace}

\newcommand{\gnoisy}{$\textsc{gen}_{\text{init}}$\xspace}
\newcommand{\gnoisyt}{$\textsc{gen}_{\text{init}}^{*}$\xspace}

\newcommand{\gours}{$\textsc{gen}_{\text{corr}}$\xspace}
\newcommand{\gcorr}{\gours}

\newcommand{\corrg}{$G$}

\newcommand{\roberta}{RoBERTa\xspace}

\newcommand{\real}[1]{\mathbb{R}^{#1}}

\newcommand{\upd}{$\mathbf{S}$\xspace}
\newcommand{\hypo}{$\mathbf{H}$\xspace}
\newcommand{\x}{$\mathbf{x}$\xspace}

\newcommand{\pre}{$\mathbf{P}$\xspace}
\newcommand{\phu}{$\mathbf{PHS}$\xspace}

\newcommand{\nodemoe}{\textbf{\textsc{moe-v}}\xspace}
\newcommand{\graphmoe}{\textbf{\textsc{moe-gx}}\xspace}
\newcommand{\atomic}{$\delta$-\textsc{atomic}\xspace}
\newcommand{\snli}{$\delta$-\textsc{snli}\xspace}
\newcommand{\social}{$\delta$-\textsc{social}\xspace}
\newcommand{\str}{\textsc{str}\xspace}
\newcommand{\gengraph}{$\mathbf{G}$\xspace}

\def\@withdot.{\ifmmode\!\string/\!
               \else\kern-1.8pt\string/\kern-1.8pt\fi.}

\newcommand{\dques}{(\pre, \hypo, \upd)\xspace}

\newcommand{\squishlist}{
  \begin{list}{$\bullet$}
    { \setlength{\itemsep}{0pt}      \setlength{\parsep}{3pt}
      \setlength{\topsep}{3pt}       \setlength{\partopsep}{0pt}
      \setlength{\leftmargin}{1.5em} \setlength{\labelwidth}{1em}
      \setlength{\labelsep}{0.5em} } }
\newcommand{\reallysquishlist}{
  \begin{list}{$\bullet$}
    { \setlength{\itemsep}{0pt}    \setlength{\parsep}{0pt}
      \setlength{\topsep}{0pt}     \setlength{\partopsep}{0pt}
      \setlength{\leftmargin}{0.2em} \setlength{\labelwidth}{0.2em}
      \setlength{\labelsep}{0.2em} } }

 \newcommand{\squishend}{
     \end{list} 
 }

\begin{document}
\maketitle

\begin{abstract}
Defeasible reasoning is the mode of reasoning where conclusions can be overturned by taking into account new evidence. Existing cognitive science literature on defeasible reasoning suggests that a person forms a \textit{mental model} of the problem scenario before answering questions. Our research goal asks whether neural models can similarly benefit from envisioning the question scenario before answering a defeasible query. Our approach is, given a question, to have a model first create a graph of relevant influences, and then leverage that graph as an additional input when answering the question. Our system, \ours, achieves a new state-of-the-art on three different defeasible reasoning datasets. This result is significant as it illustrates that performance can be improved by guiding a system to ``think about'' a question and explicitly model the scenario, rather than answering reflexively.\footnote{Code and data located at \url{https://github.com/madaan/thinkaboutit}}

\end{abstract}

\section{Introduction \label{introduction}}


Defeasible inference is a mode of reasoning where additional information can modify conclusions \cite{sep-reasoning-defeasible}.
Here we consider the specific formulation and challenge in \citet{rudinger-etal-2020-thinking}:
Given that some premise \pre plausibly implies a hypothesis \hypo, does
new information that the situation is \upd weaken or strengthen the conclusion \hypo?
For example, consider the premise ``The drinking glass fell'' with a possible implication ``The glass broke''.
New information that ``The glass fell on a pillow'' here {\it weakens} the implication.

\begin{figure}[!ht]
\centering
{\includegraphics[width=0.78\columnwidth]{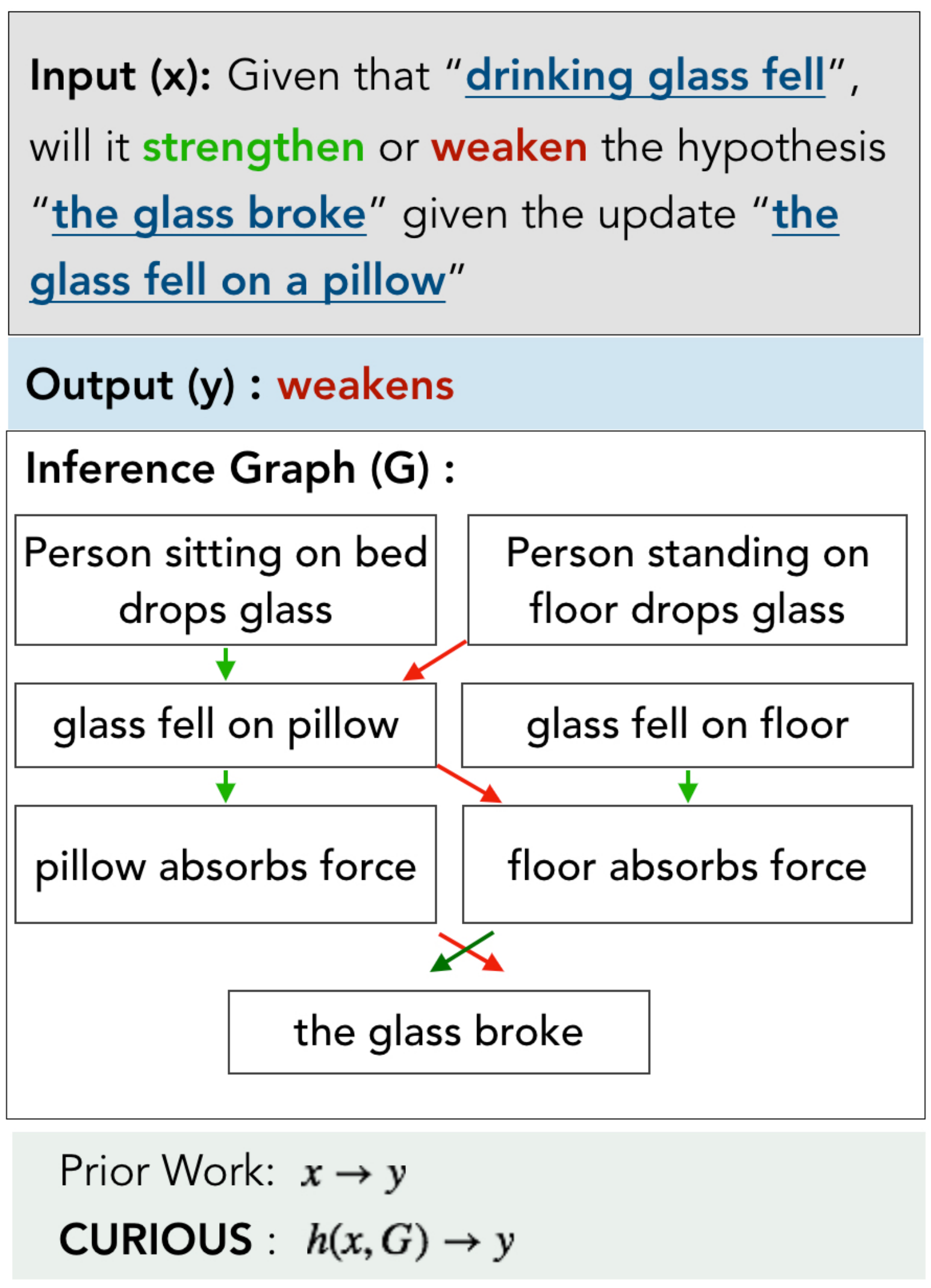}}
\caption{\ours improves defeasible reasoning by modeling the question scenario with an inference graph $G$ adapted from cognitive science literature. The graph is encoded judiciously using our graph encoder $h(.)$, resulting in end task performance improvement.}
\label{fig:deftrain}
\end{figure}

We borrow ideas from the cognitive science literature that supports defeasible reasoning for humans with an \emph{inference graph} \citep{Pollock2009ARS,Pollock1987DefeasibleR}.  Inference graphs formulation in \cite{defeasible-human-helps-paper}, which we use in this paper, draws connections between the \pre, \hypo, and \upd through mediating events. This can be seen as a \emph{mental model} of the question scenario before answering the question \cite{JohnsonLaird1983MentalM}. 
This paper asks the natural question: can modeling the question scenario with inference graphs help machines in defeasible reasoning? 

\begin{figure}[!t]
\centering
{\includegraphics[width=\columnwidth]{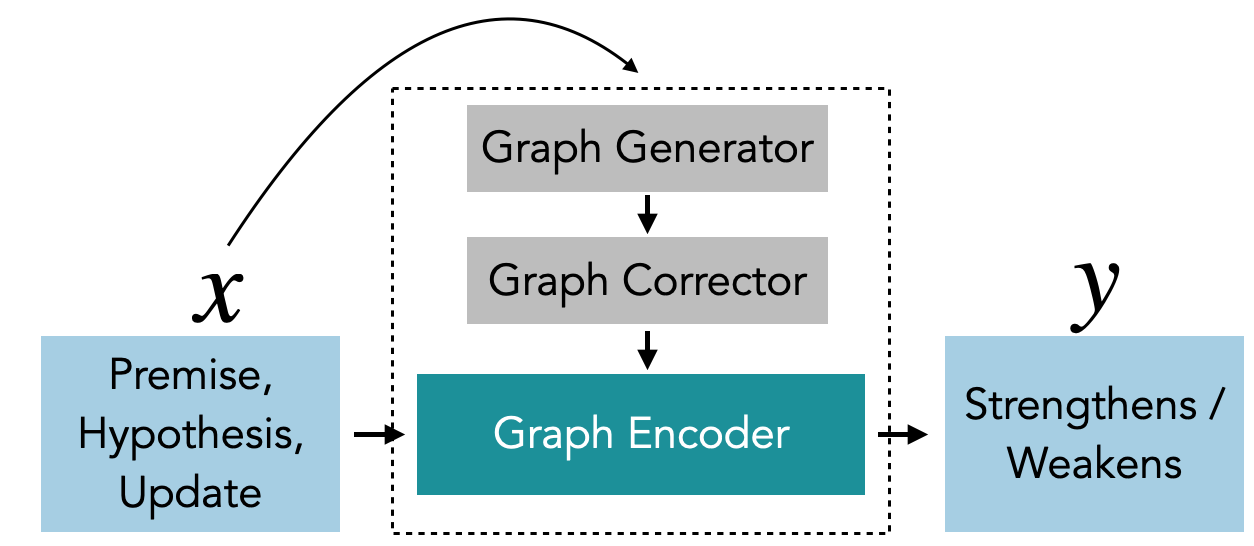}}
\caption{An overview of \ours} \label{fig:overview}
\end{figure}

Our approach is as follows. First, given a question, generate an inference graph describing important influences between question elements. Then, use that graph as an additional input when answering the defeasible reasoning query. Our proposed system, \ours, comprises a graph generation module and a graph encoding module to use the generated graph for the query~(Figure ~\ref{fig:overview}).

To generate inference graphs, we build upon past work that uses a sequence to sequence approach \cite{defeasible-human-helps-paper}. 
However, our analysis revealed that the graphs can often be erroneous, and \ours also includes an error correction module to generate higher quality inference graphs. This was important because we found that better graphs are more helpful in the downstream QA task.

The generated inference graph is then used for the QA task on three existing defeasible inference datasets from diverse domains, viz., \snli (natural language inference) \cite{bowman2015large}, \social (reasoning about social norms) \cite{forbes2020social}, and \atomic (commonsense reasoning) \cite{sap2019atomic}. We show that the way the graph is encoded for input is important.
If we simply augment the question with the generated graphs, there are some gains on all datasets. However, the accuracy improves substantially across all datasets with a more judicious encoding of the graph-augmented question that accounts for interactions between the graph nodes. To achieve this, we use the mixture of experts approach \cite{jacobs1991adaptive} to include a mixture of experts layers during encoding, enabling the ability to attend to specific nodes while capturing their interactions selectively. 

In summary, our contribution is in drawing on the idea of an inference graph from cognitive science to show benefits in a defeasible inference QA task. Using an error correction module in the graph generation process, and a judicious encoding of the graph augmented question, \ours achieves a new state-of-the-art over three defeasible datasets.
This result is significant also because our work illustrates that guiding a system to ``think about" a question before answering can improve performance.

\begin{figure*}[!ht]
\centering
{\includegraphics[width=0.93\textwidth,height=0.25\textheight]{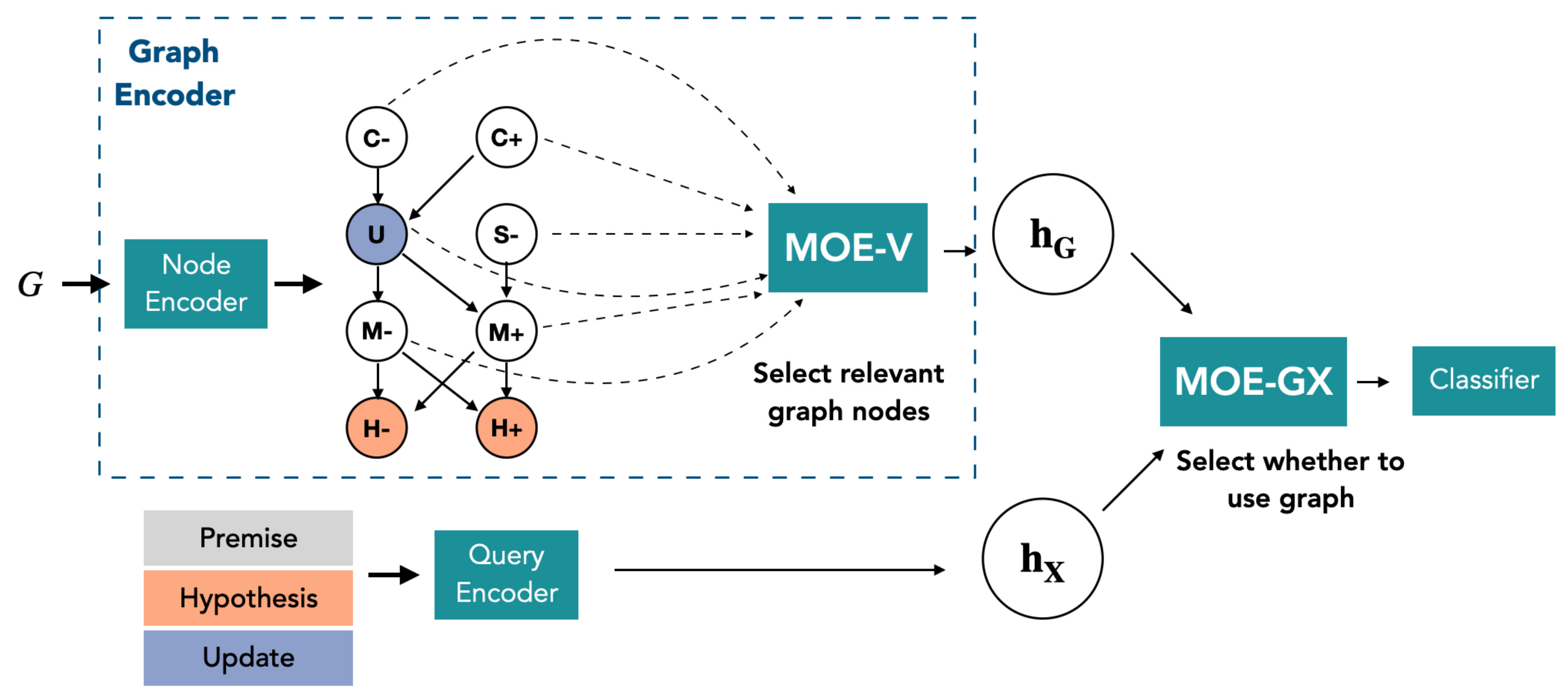}}
\caption{An overview of our method to perform graph-augmented defeasible reasoning using a hierarchical mixture of experts. First, \nodemoe selectively pools the node representations to generate a representation \hg of the inference graph. Then, \graphmoe pools the query representation \hq and the graph representation generated by \nodemoe to pass to the upstream classifier.}
\label{fig:moe}
\end{figure*}

\section{Task}

\label{sec:task-defeasible}
We use the defeasible inference task and datasets defined in \cite{rudinger-etal-2020-thinking}, namely given an input \x= (\pre,\hypo,\upd), predict the output $\V{y} \in \{strengthens, weakens\}$, where \pre, \hypo, and \upd{} are sentences describing a premise, hypothesis, and scenario respectively, and $y$ denotes whether $S$ strengthens/weakens the plausible conclusion that \hypo{} follows from \pre, as described in Section~\ref{introduction}.

\eat{
Defeasible inference~\cite{rudinger-etal-2020-thinking} is a mode of reasoning in which given a premise \pre, a hypothesis \hypo may be strengthened or weakened in light of new evidence \upd.
For example, given a premise \textit{The drinking glass fell}, the hypothesis \textit{The glass broke} will be weakened by the situation \textit{the glass fell on the pillow}, and strengthened by the situation \upd \textit{the glass fell on the rocks}. We define the input \x= (\pre,\hypo,\upd) and output $\V{y} \in \{strengthens, weakens\}$.
}

\section{Approach}

Inspired by past results \cite{defeasible-human-helps-paper} that humans found inference graphs useful for defeasible inference,  we investigate whether neural models can benefit from envisioning the question scenario using an inference graph before answering a defeasible inference query.

\paragraph{Inference graphs} As inference graphs are central to our work, we give a brief description of their structure next.
Inference graphs were introduced in philosophy by \citet{Pollock2009ARS} to aid defeasible reasoning for humans, and in NLP by \citet{tandon2019wiqa} for a counterfactual reasoning task. 
We interpret the inference graphs as having four kinds of nodes~\cite{Pollock2009ARS,defeasible-human-helps-paper}: 
\squishlist
\item[i.] \textbf{Contextualizers (C-, C+):} these nodes set the context around a situation and connect to the \pre.
\item[ii.] \textbf{Situations (S, S-):} these nodes are new situations that emerge which might overturn an inference.
\item[iii.] \textbf{Hypothesis (H-, H+):} Hypothesis nodes describe the outcome/conclusion of the situation.
\item[iv.] \textbf{Mediators (M-, M+):} Mediators are nodes that help bridge the knowledge gap between a situation and a hypothesis node by explaining their connection explicitly. These node can either act as a \emph{weakener} or \emph{strengthener}.
\squishend
Each node in an influence graph is labeled with an event (a sentence or a phrase). The signs \textbf{-} and \textbf{+} capture the nature of the influence event node. Concrete examples are present in Figures~\ref{fig:deftrain}, \ref{fig:defeasible_incorrect_example}, and in Appendix~\secref{sec:igraphschema}.

\subsection{Overview of \ours}

Our system, \ours, comprises three components, (i) a graph generator \gnoisy, (ii) a graph corrector \gcorr, (iii) a graph encoder (Figure~\ref{fig:deftrain}). 
\gnoisy generates an inference graph from the input $\V{x}$. We borrow the sequence to sequence approach of \gnoisy from \citet{defeasible-human-helps-paper} without any architectural changes. However, we found that the resulting graphs can often be erroneous (which hurts task performance), so \ours includes an error correction module \gcorr to generate higher quality inference graphs that are then judiciously encoded using the graph encoder. This encoded representation is then passed through a classifier to generate an end task label.
The overall architecture is shown in Figure \ref{fig:overview}.

\subsection{Graph generator}
\label{sec:graphgen}
\begin{figure}[!htb]
  \includegraphics[width=\linewidth]{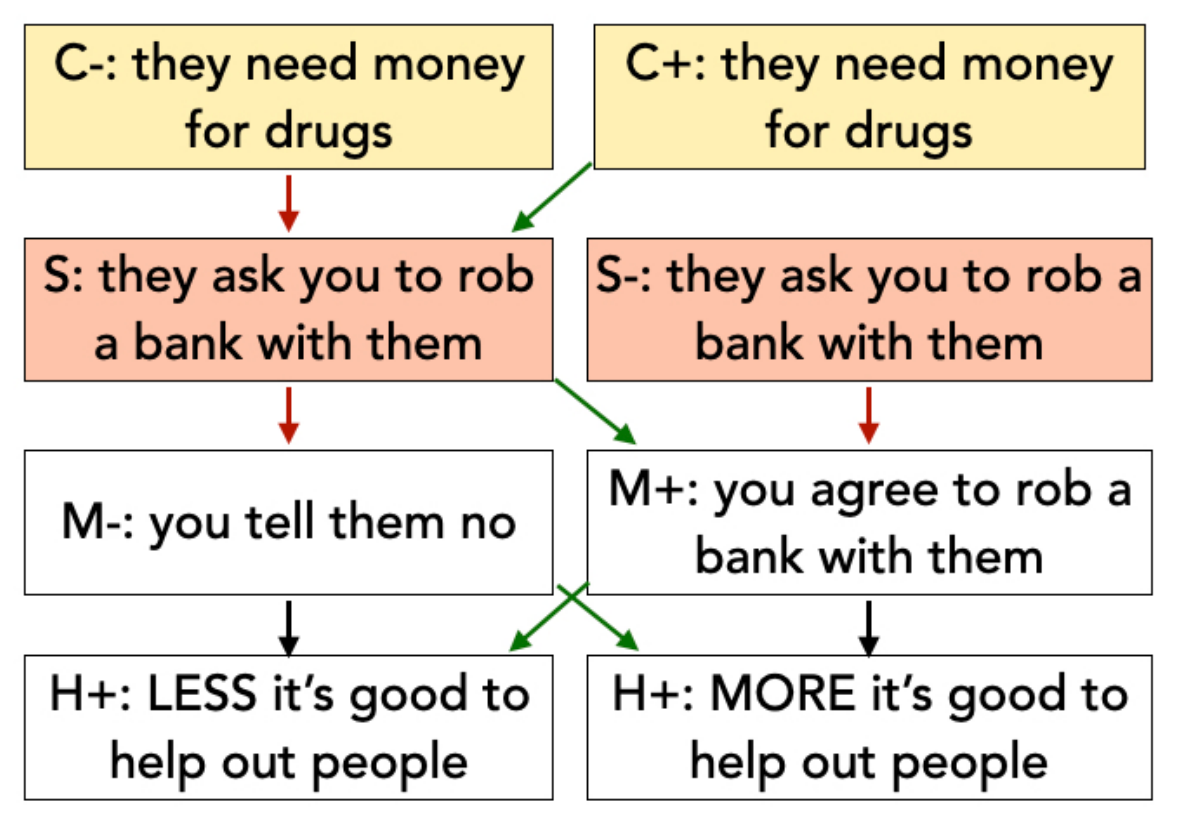}
\caption{The graphs generated by \gnoisy.The input graph has repetitions for nodes $\{C{-}, C{+}\}$ and  $\{S, S{-}\}$. The corrected graph generated by \gcorr replaces the repetitions with meaningful labels.}
\label{fig:defeasible_incorrect_example}
\end{figure}

As the initial graph generator, we use the method described in \citet{defeasible-human-helps-paper}  (\gnoisy) to generate inference graphs for defeasible reasoning.\footnote{We use their publicly available code and data}
Their approach involves first training a graph-generation module, and then performing zero-shot inference on a defeasible query to obtain an inference graph.
They obtain training data for the graph-generation module from \wiqa dataset~\cite{tandon2019wiqa}.
\wiqa is a dataset of 2107 $(\V{T}_i, \V{G}_i)$ tuples, where $\V{T}_i$ is the passage text that describes a process~(e.g., waves hitting a beach), and $\V{G}_i$ is the corresponding influence graph.
The graph generator \gnoisy is trained as a seq2seq model, by setting $\texttt{input} = \text{[Premise] } \V{T}_i \mid \text{[Situation] } \V{S}_i \mid \text{[Hypothesis] } \V{H}_i$, and $\texttt{output} = \V{G}_i$.
Note that $\V{S}_i$ and $\V{H}_i$ are nodes in the influence graph $\V{G}_i$, allowing grounded generation.
$\text{[Premise], [Situation], [Hypothesis]}$ are special tokens used to demarcate the input.

\subsection{Graph corrector}
\label{sec:graphcorrector}
We found that 70\% of the randomly sampled 100 graphs produced by \gnoisy 
(undesirably) had \textit{repeated} nodes (an example of repeated nodes is in Figure~\ref{fig:defeasible_incorrect_example}).
Repeated nodes introduce noise because they violate the semantic structure of a graph, e.g., in Figure~\ref{fig:defeasible_incorrect_example}, nodes C+ and C- 
are repeated, although they are expected to have opposite semantics. 
Higher graph quality yields better end task performance when using inference graphs (as we will show in \S\ref{sec:graph_corrector})

To repair such problems, we train a graph corrector, \gours, that takes as input \badg, and as output it gives a graph \betterg, with repetitions fixed. To train the model, we require (\badg, \betterg) examples, which we generate using a data augmentation technique described in the Appendix~\secref{sec:trainingdatacorrector}. Because the nodes in the graph are from an open vocabulary, we then train a T5 sequence-to-sequence model \cite{raffel2020exploring} with input = \badg and output = \betterg.
In summary, given a defeasible query \phu, we generate a potentially incorrect initial graph \badg using \gnoisy.
We then feed \badg to \gcorr to obtain an improved graph $\V{G}$.

\subsection{Graph Encoder}

For each defeasible query \dques, we add the inference graph $\V{G}$ from \ours (the corrected graph from \secref{sec:graphcorrector}), to provide additional context for the query, as we now describe.

We concatenate the components \dques of the defeasible query into a single sequence of tokens $\V{x} = (\mathbf{P} \| \mathbf{H} \| \mathbf{S})$, where $\|$ denotes concatenation.
Thus, each sample of our graph-augmented binary-classification task takes the form $((\V{x}, \V{G}), \V{y})$, where  $\V{y} \in \{\text {\textit{strengthener}, \textit{weakener}}\}$.
Following~\cite{defeasible-human-helps-paper}, we do not use edge labels and treat all the graphs as undirected graphs.

\paragraph{Overview:} We first use a language model \lm to obtain a dense representation \hq for the defeasible query \x, and a dense representation \hv for each node $\V{v} \in \V{G}$. The node representations \hv are then pooled using a hierarchical mixture of experts (\moe) to obtain a graph representation \hg.
The query representation \hq and the graph representation \hg are combined to solve the defeasible task.
We now provide details on obtaining \hq, \hv, \hg.

\subsubsection{Encoding the query and nodes}

Let \lm be a pre-trained language model (in our case \roberta~\cite{liu2019roberta}).
We use $\V{h}_{\V{S}} = \mathcal{L}(\V{S}) \in \real{d}$ to denote the dense representation of sequence of tokens $\V{S}$ returned by the language model \lm.
Specifically, we use the pooled representation of the beginning-of-sequence token \texttt{<s>} as the sequence representation. 

We encode the defeasible query $\V{x}$ and the nodes of the graph using \lm. Query representation is computed as $\V{h}_\V{x} = \C{L}(\V{x})$, and we similarly obtain a matrix of node representations $\nodeset$ by encoding each node $\V{v}$ in \gengraph with \lm as follows: 
\begin{align}
    \V{h}_{\V{V}} = [\V{h}_{v_1}; \V{h}_{v_2}; \ldots; \V{h}_{|\V{V}|}]
    \label{eqn:encode_nodes}
\end{align}

where $\V{h}_{v_i} \in \real{d}$ refers to the dense representation for the $i^{th}$ node of $\V{G}$ derived from \lm (i.e., $\V{h}_{v_i} = \C{L}(v_i)$), and $\V{h}_{\V{V}} \in \real{|\V{V}| \times d}$ to refer to the matrix of node representations.

\subsubsection{Graph representations using \moe}
\label{sec:moe_pooling}

Recently, mixture-of-experts~\cite{jacobs1991adaptive,shazeer2017outrageously,fedus2021switch} has emerged as a promising method of combining multiple feature types.
Mixture-of-experts (\moe) is especially useful when the input consists of multiple \textit{facets}, where each facet has a specific semantic meaning.
Previously, \citet{gu2018universal,chen2019multi} have used the ability of \moe to pool disparate features on low-resource and cross-lingual language tasks.
Since each node in the inference graphs used by us plays a specific role in defeasible reasoning (contextualizer, situation node, or mediator), 
we take inspiration from these works to design a hierarchical \moe model~\cite{jordan1995convergence} to pool node representations $\nodeset$ into a graph representation \hg.

An \moe consists of $n$ expert networks $\V{E_1}, \V{E_2}, \ldots, \V{E_n}$ and a gating network $\V{M}$.
Given an input $\V{x} \in \real{d}$, each expert network $\V{E_i}: \real{d} \rightarrow \real{k}$ learns a transform over the input.
The gating network $\V{M}: \real{d} \rightarrow \Delta^d$ gives the weights $\V{p} = \{p_1, p_2, \ldots, p_n\}$ to combine the expert outputs for input $\V{x}$.
Finally, the output $\V{y}$ is returned as a convex combination of the expert outputs:
\vspace{-0.5em}
\begin{align}
    \V{p} &= \V{M}(\V{x}) \nonumber \\
     \V{y} &= \sum_{i=1}^{n} p_i \V{E_i(x)}
    \label{eqn:moe}
\end{align}

The output $\V{y}$ can either be the logits for an end task~\cite{shazeer2017outrageously,fedus2021switch} or pooled features that are passed to a downstream learner~\cite{chen2019multi,gu2018universal}.
The gating network $\V{M}$ and the expert networks $\V{E_1}, \V{E_2}, \ldots, \V{E_n}$ are trained end-to-end.
During learning, the gradients to $\V{M}$ train it to generate a distribution over the experts that favors the best expert for a given input.
Appendix~\secref{sec:moegradientanalysis} presents a further discussion on our \moe formulation and an analysis of the gradients.

\paragraph{Hierarchical \moe for defeasible reasoning}

Different parts of the inference graphs might help answer a query to a different degree.
Further, for certain queries, graphs might not be helpful (and could even be distracting), and the model could rely primarily on the input query alone.
This motivates a two-level architecture that can: (i) select a subset of the nodes in the graph and ii) selectively reason across the query and the graph to varying degrees.

Given these requirements, a hierarchical \moe~\cite{jordan1994hierarchical} model presents itself as a natural choice to model this task.
The first \moe (\nodemoe) creates a graph representation by taking a convex combination of the node representations.
The second \moe (\graphmoe) then takes a convex-combination of the graph representation returned by \nodemoe and query representation and passes it to an MLP for the downstream task.

\squishlist
\item  \nodemoe consists of five node-experts and gating network to selectively pool node representations \hv to graph representation \hg:
\vspace{-0.1em}
\begin{align}
    \V{p} &= \V{M}(\V{h}_\V{V}) \nonumber \\
     \V{h}_\V{G} &= \sum_{v \in \V{V}} p_v E_v(v) 
    \label{eqn:moegraph}
\end{align}

\item \graphmoe contains two experts (graph expert $E_\V{G}$ and question expert $E_\V{Q}$) and a gating network to combine the graph representation \hg returned by \graphmoe and the query representation \hq:
\vspace{-0.2em}
\begin{align}
    \V{p} &= \V{M}([ \V{h}_\V{G}; \V{h}_\V{Q} ]) \nonumber \\
     \V{h}_\V{y} &= E_\V{G}(\V{h}_\V{G}) + E_\V{Q}(\V{h}_\V{Q})
    \label{eqn:moenodequery}
\end{align}

\squishend

$\V{h}_\V{y}$ is then passed to a 1-layer MLP to perform classification.
The gates and the experts in our \moe model are single-layer MLPs, with equal input and output dimensions for the experts.

\section{Experiments}
In this section, we empirically investigate if \ours can improve defeasible inference by first modeling the question scenario using inference graphs. We also study the reasons for any improvements.

\subsection{Experimental setup}

\begin{table}[!ht]
\centering
\begin{tabular}{llrr}
\toprule
Dataset & Split & \# Samples & Total\\ \midrule
 \multirow{3}{*}{\atomic} & train & 35,001 & \multirow{3}{*}{42,977} \\
      & test   & 4137  \\
& dev  & 3839 \\ \midrule
 \multirow{3}{*}{\social} & train & 88,675 & \multirow{3}{*}{92,295} \\
      & test   & 1836  \\
& dev  & 1784 \\ \midrule
 \multirow{3}{*}{\snli} & train & 77,015 & \multirow{3}{*}{95,795} \\
      & test   & 9438  \\
& dev  & 9342 \\
\bottomrule
\end{tabular}
\caption{Number of samples in each dataset by split. 
}
\label{table:data-split}
\end{table}

\paragraph{Datasets}
Our end task performance is measured on the three existing datasets for defeasible inference
provided by \citet{rudinger-etal-2020-thinking}:\footnote{ \url{github.com/rudinger/defeasible-nli}} \atomic, \snli, \social (Table~\ref{table:data-split}).
These datasets exhibit substantial diversity because of their domains: \snli (natural language inference), \social (reasoning about social norms), and \atomic (commonsense reasoning). Thus, it would require a general model to perform well across these diverse datasets.

\paragraph{Baselines and setup}
The previous state-of-the-art (SOTA) is the \roberta~\cite{liu2019roberta} model presented in \citet{rudinger-etal-2020-thinking}, and we report the published numbers for this baseline. 
For an additional baseline, we directly use the initial inference graph \badg{} generated by \gnoisy,
and provide it to the model simply as a string (i.e., sequence of tokens; a simple, often-used approach).
This baseline is called \textsc{e2e}-STR. We use the same hyperparameters as \citet{rudinger-etal-2020-thinking}, and add a detailed description of the hyperparameters in Appendix~\secref{sec:hyperparams}.
For all the \qa experiments, we report the accuracy on the test set using the checkpoint with the highest accuracy on the development set.
We use the McNemar's test~\cite{mcnemar1947note,dror2018hitchhiker} and use $p < 0.05$ as a threshold for statistical significance.
All the p-values are reported in Appendix~\secref{sec:statsig}.

\subsection{Results}
\label{sec:rq2}
Table \ref{tab:overall-main-result} compares QA accuracy on these datasets without and with modeling the question scenario. The results suggest that we get consistent gains across all datasets, with \snli gaining about 4 points. \ours achieves a new state-of-the-art across three datasets, as well as now producing justifications for its answers with inference graphs.

\newcommand{\tabw}{1.61cm}
\begin{table}[!ht]
\centering
\begin{tabular}{p{1.59cm}p{\tabw}p{\tabw}p{\tabw}p{\tabw}}
\toprule
 & \atomic & \snli & \social \\
\midrule
\small{Prev-SOTA}   & 78.3\    & 81.6\  & 86.2\   \\
\small{\textsc{e2e}-STR}  & 78.8\    & 82.2\  & 86.7\ \\
\ours &   \textbf{80.2*}\   & \textbf{85.6*}\  & \textbf{88.6*}\ \\
\bottomrule
\end{tabular}
\caption{\ours is better across all the datasets. This demonstrates that understanding the question scenario through generating an inference graph helps. * indicates statistical significance.}
\label{tab:overall-main-result}
\end{table}

\label{sec:results}

\subsection{Understanding \ours gains}
In this section, we study the contribution of the main components of the \ours pipeline. 

\subsubsection{Impact of graph corrector}
\label{sec:graph_corrector}

We ablate the graph corrector module \gcorr in \ours by directly supplying the output from \geninit to the graph encoder. Table~\ref{tab:better_graphs_better_perf} shows that this ablation consistently hurts across all the datasets. \gcorr provides 2 points gain across datasets. This indicates that better graphs lead to better task performance, assuming that \gcorr actually reduces the noise. Next, we investigate if \gcorr can produce more informative graphs.

\begin{table}[!ht]
\centering
\begin{tabular}{lrrr}
\toprule
 & \atomic & \snli & \social \\
\midrule
\badg  & 78.5  & 83.8 & 88.2   \\
\corrg &   \textbf{80.2}*  & \textbf{85.6}* & \textbf{88.6}   \\
\bottomrule
\end{tabular}
\caption{Performance w.r.t. the graph used. \badg{} is the initial graph from \geninit, \corrg{} is the corrected graph from \gcorr.
Better graphs lead to better task performance. * indicates statistical significance.}
\label{tab:better_graphs_better_perf}
\end{table}

\paragraph{Do graphs corrected by \gcorr show fewer repetitions?}
We evaluate the repetitions in the graphs produced by \geninit and \gcorr using two metrics: (i) repetitions per graph: average number of repeated nodes in a graph. (ii) \% with repetitions: \% of graphs with at least one repeated node.

\begin{table}[!ht]
\centering
\begin{tabular}{ll|ll}
\hline
        & Repetitions & \geninit & \gcorr \\ \hline
\atomic & per graph            & 2.05 & \textbf{1.26}   \\ \cline{2-4} 
        & \% graphs            & 72   & \textbf{48 }    \\ \hline
\snli   & per graph            & 2.09 & \textbf{1.18}   \\ \cline{2-4} 
        & \% graphs            & 73   & \textbf{46}     \\ \hline
\social & per graph            & 2.2  & \textbf{1.32}   \\ \cline{2-4} 
        & \% graphs            & 75   & \textbf{49}     \\ \hline
\textsc{overall} & per graph   &      & $\Delta$ \textbf{-40\%}   \\ \cline{2-4} 
        & \% graphs            &      & $\Delta$ \textbf{-25.7\%}     \\ \hline
\end{tabular}
\caption{\gcorr reduces the inconsistencies in graphs. The number of repetitions per graph and percentage of graphs with some repetition, both improve.}
\label{tab:graph-gcorr-noise-reduces}
\end{table}
 
Table~\ref{tab:graph-gcorr-noise-reduces} shows \gcorr does reduce repetitions by approximately 40\% (2.11 to 1.25) per graph across all datasets, and also reduces the fraction of graphs with at least one repetition by $~$25.7\% across. 

\subsubsection{Impact of graph encoder}
\label{sec:graph_encoding_gains}

We experiment with two alternative approaches to graph encoding to compare our \moe approach by using the graphs generated by \gcorr:

\noindent\textbf{1. Graph convolutional networks}: We follow the approach of \citet{lv2020graph} who use  \gcn~\cite{kipf2016semi} to learn rich node representations from graphs. Broadly, node representations are initialized by \lm and then refined using a \gcn. Finally, multi-headed attention~\cite{vaswani2017attention} between question representation \hq and the node representations is used to yield \hg. We add a detailed description of this method in Appendix~\secref{sec:gcn_pooling}.

\noindent\textbf{2. String based representation}: Another popular approach \cite{proscript} is to concatenate the string representation of the nodes, and then using \lm to obtain the graph representation \hg$=\C{L}(v_1 \| v_2 \| .. )$ where $\|$ denotes string concatenation.

Table~\ref{tab:better_encoding_better_perf_1}  shows that \moe graph encoder improves end task performance significantly compared to the baseline.\footnote{Appendix~\secref{sec:runtime} provides an analysis on time and memory requirements.}
In the following analysis, we study the reasons for these gains in-depth. 

We hypothesize that \gcn is less resistant to noise than \moe in our setting, thus causing a lower performance.
The graphs augmented with each query are not human-curated and are instead generated by a language model in a zero-shot inference setting. Thus, the \gcn style message passing might amplify the noise in graph representations. 
On the other hand, \nodemoe first selects the most useful nodes to answer the query to form the graph representation \hg.
Further, \graphmoe can also decide to completely discard the graph representations, as it does in many cases where the true answer for the defeasible query is \textit{weakens}.

To further establish the possibility of message passing hampering the downstream task performance, we experiment with a \gcn-\moe hybrid, wherein we first refine the node representations using a 2-layer \gcn as used by \cite{lv2020graph}, and then pool the node representations using an \moe.
We found the results to be about the same as ones we obtained with \gcn~(3rd-row Table~\ref{tab:better_encoding_better_perf_1}),  indicating that bad node representations are indeed the root cause for the bad performance of \gcn. This is also supported by  \citet{Shi2019FeatureAttentionGCNUnderNoise} who found that noise propagation directly deteriorates network embedding and \gcn is sensitive to noise.

Interestingly, graphs help the end-task even when encoded using a relatively simple \str based encoding scheme, further establishing their utility.

\begin{table}[!ht]
\centering
\begin{tabular}{lrrr}
\toprule
 & \atomic & \snli & \social \\
\midrule
\str & 79.5 & 83.1 & 87.2  \\
\gcn & 78.9 & 82.4 & 88.1  \\
\gcn + \moe & 78.7 & 84.3 & 87.8 \\
\moe  &  \textbf{80.2} & \textbf{85.6} & \textbf{88.6}  \\
\bottomrule
\end{tabular}
\caption{Contribution of MoE-based graph encoding compared with alternative graph encoding methods. The gains of \moe over \gcn are statistically significant for all the datasets, and the gains over \str are significant for \snli and \social.}
\label{tab:better_encoding_better_perf_1}
\end{table}

\subsubsection{Detailed \moe analysis}
\label{sec:detailedmoeanalysis}

We now analyze the two MoEs used in \ours: (i) the \moe over the nodes (\nodemoe), and (ii) the \moe over $\V{G}$ and input $x$ (\graphmoe). 

\begin{figure}[!ht]
\centering
{\includegraphics[width=0.40\textwidth,height=0.22\textheight]{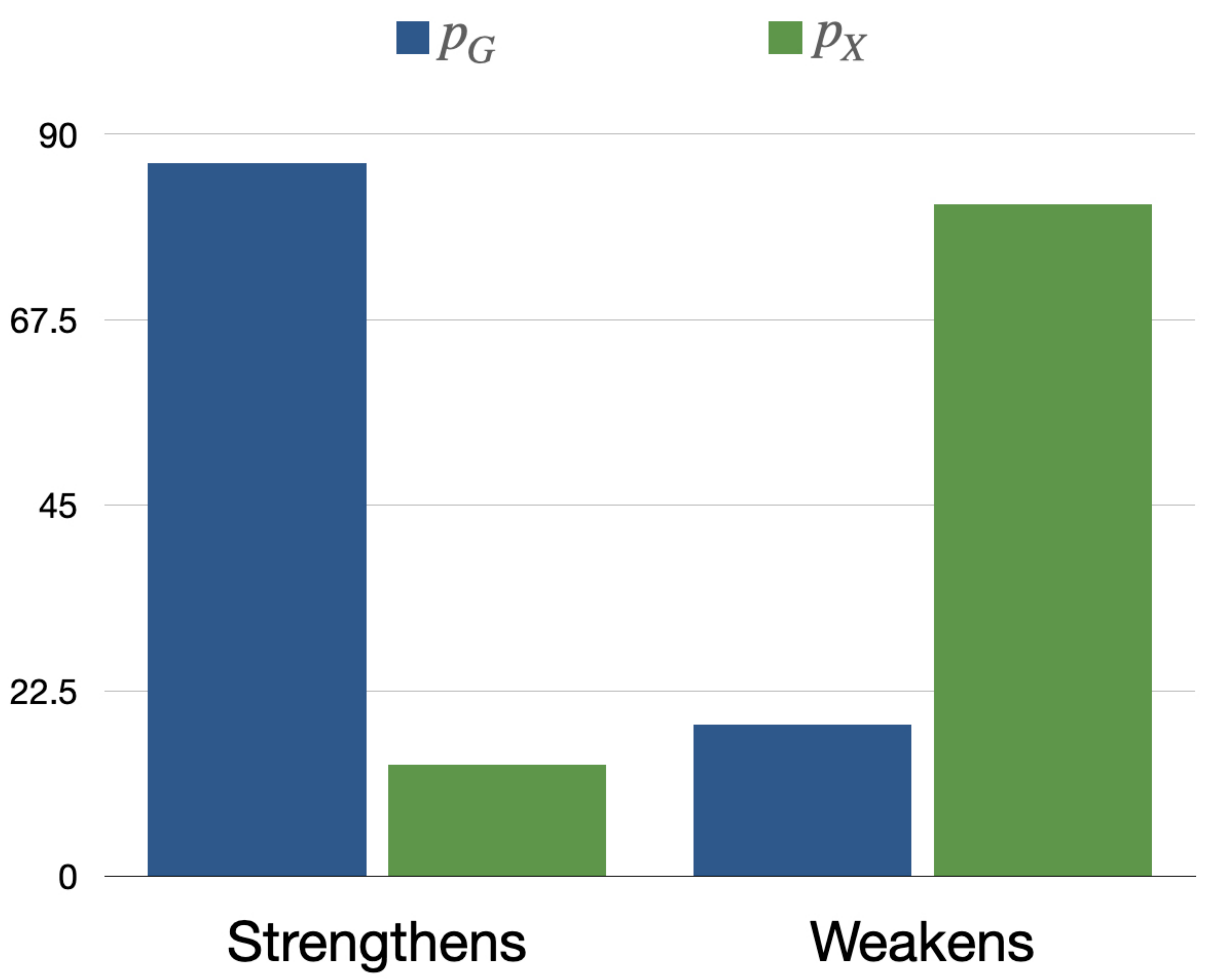}}
\caption{\graphmoe gate values for the classes strengthens and weakens, averaged over the three datasets.}
\label{fig:graphquesmoedist}
\end{figure}

\paragraph{\graphmoe performs better for $y=\text{strengthens:}$} Figure~\ref{fig:graphquesmoedist} shows that the graph makes a stronger contribution than the input, when the label is \textit{strengthens}. 
In instances where the label is \textit{weakens}, the gate of \graphmoe gives a higher weight to the question. 
This trend was present across all the datasets. We conjecture that this happens because language models are tuned to generate events that happen rather than events that do not. In the case of a weakener, the nodes must be of the type \textit{event1 leads to less of event2}, whereas language models are naturally trained for \textit{event1 leads to event2}. 
Understanding this in-depth requires further investigation in the future. 

\begin{figure}[!ht]
\centering
{\includegraphics[width=0.48\textwidth]{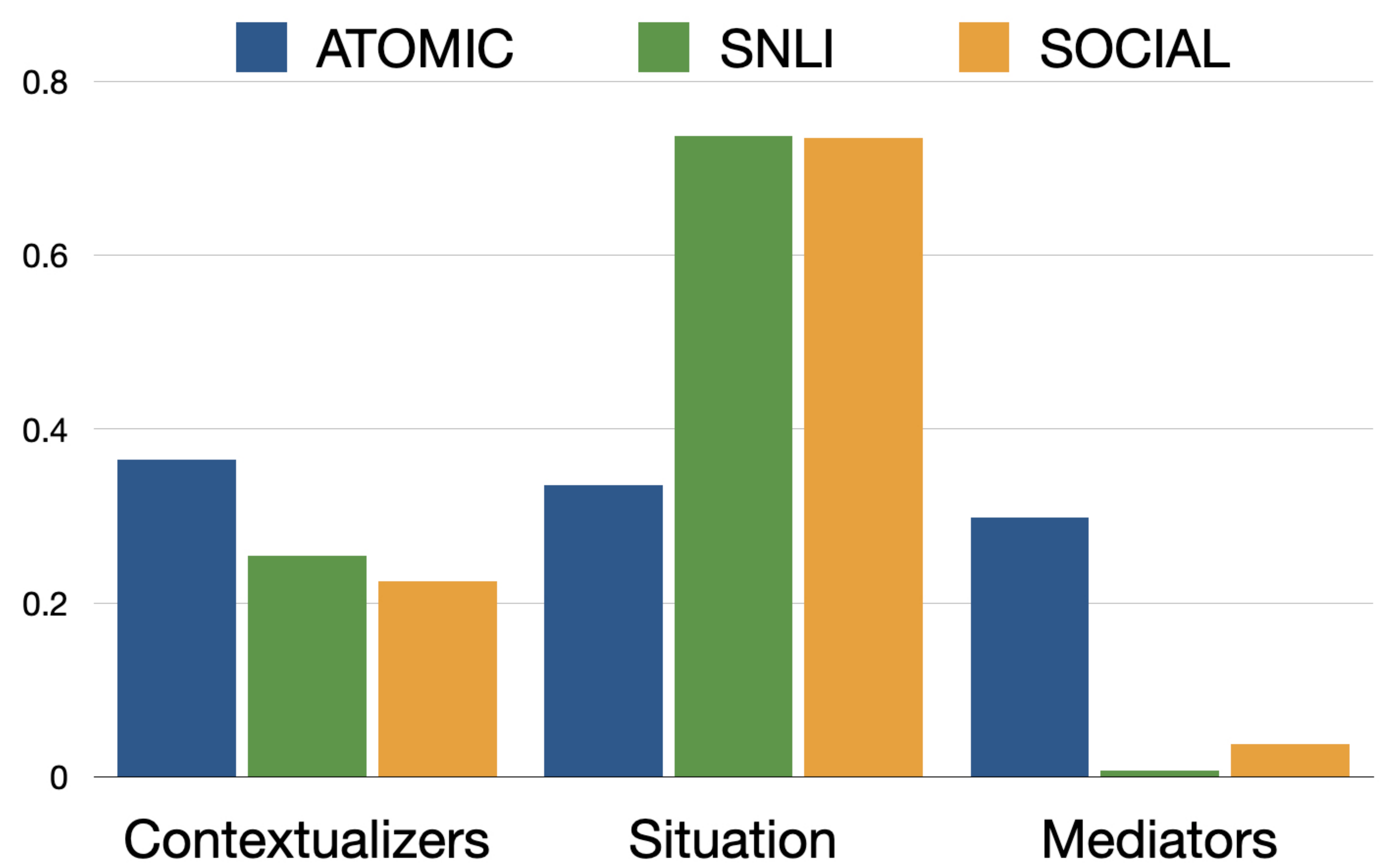}}
\caption{\nodemoe gate values for the three datasets.}
\label{fig:nodemoedist}
\vspace{-4mm}
\end{figure}

\paragraph{\nodemoe relies more on specific nodes:} We study the distribution over the types of nodes and their contribution to \nodemoe. Recall from Figure \ref{fig:moe} that C- and C+ nodes are contextualizers that provide more background context to the question, and S- node is typically an inverse situation (i.e., inverse \upd), while M- and M+ are the mediator nodes leading to the hypothesis. Figure \ref{fig:nodemoedist} shows that the situation node S- was the most important, followed by the contextualizer and the mediator.
Notably, our analysis shows that mediators are less important for machines than they were for humans in the experiments conducted by~\citet{defeasible-human-helps-paper}. This is probably because humans and machines use different pieces of information. As our error analysis shows in \S{\ref{sec:analysis}}, the mediators can be redundant given the query $\V{x}$. Humans might have used the redundancy to reinforce their beliefs, whereas machines leverage the unique signals present in S- and the contextualizers. 

\paragraph{\nodemoe, \graphmoe have a peaky distribution:} A peaky distribution over the gate values implies that the network is judiciously selecting the right expert for a given input. We compute the average entropy of \nodemoe and \graphmoe and found the entropy values to be 0.52~(max 1.61) for \nodemoe, and 0.08~(max 0.69) for \graphmoe.
The distribution of the gate values of \nodemoe is relatively flat, indicating that specialization of the node experts might have some room for improvement~(additional discussion in Appendix~\secref{sec:moegradientanalysis}).
Analogous to scene graphs-based explanations in visual QA \cite{Ghosh2019VQAExplanations}, peaky distributions over nodes can be considered as an explanation through supporting nodes.

\begin{figure}[!ht]
\centering
{\includegraphics[width=0.45\textwidth,height=0.20\textheight]{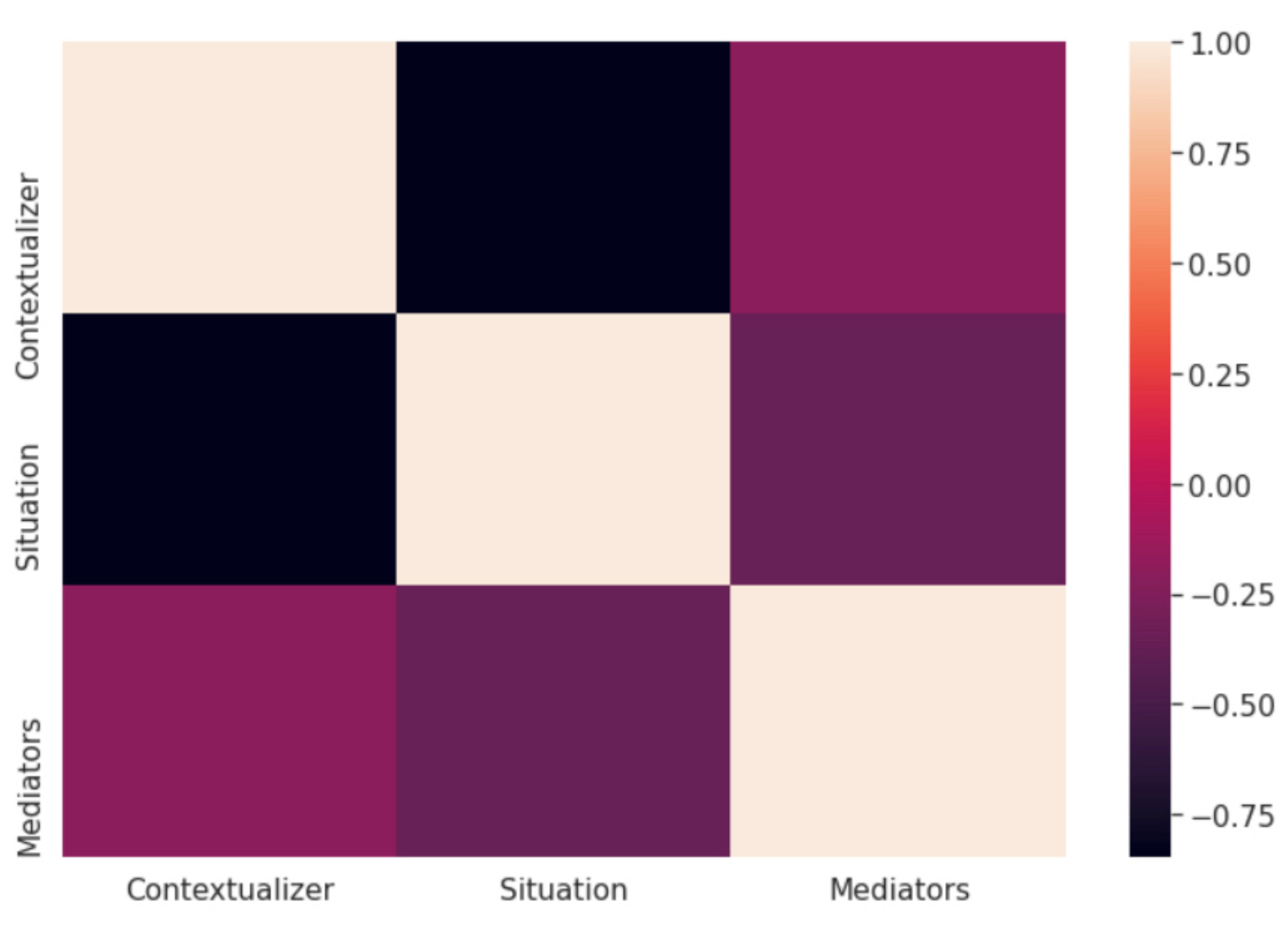}}
\caption{Correlation between probability assigned to each semantic type of the node by \nodemoe}
\label{fig:gatecorr}
\end{figure}

\paragraph{\nodemoe learns the node semantics:}  The network learned the semantic grouping of the nodes (contextualizers, situation, mediators), which became evident when plotting the correlation between the gate weights. As Figure~\ref{fig:gatecorr} shows, there is a strong negative correlation between the situation nodes and the context nodes, indicating that only one of them is activated at a time.

\section{Error analysis}
\label{sec:analysis}
\begin{table}[!ht]
\noindent
\setlength\tabcolsep{0pt}
\begin{tabular}{c >{\bfseries}r @{\hspace{0.6em}}c @{\hspace{0.4em}}c}
  \multirow{10}{*}{\parbox{1.0cm}{\bfseries\raggedleft}} & 
    & \multicolumn{2}{c}{\bfseries} \\
  & & \bfseries now fail & \bfseries now succ  \\
  & prev. fail & \ABox{\scriptsize{\atomic} \small{615}}{\scriptsize{\snli} \small{197} \\ \scriptsize{\social} \small{772}} & \ABox{\scriptsize{\atomic} \small{294}}{\scriptsize{\snli} \small{124} \\ \scriptsize{\social} \small{398}} \\[2.2em] 
  & prev. succ & \ABox{\scriptsize{\atomic} \small{207}}{\scriptsize{\snli} \small{68} \\ \scriptsize{\social} \small{302}} & 
  \ABox{\scriptsize{\atomic} \small{3022}}{\scriptsize{\snli} \small{1448} \\ \scriptsize{\social} \small{7967}}  \\
\end{tabular}
\caption{Confusion matrix: can \ours fix previously failing or successful examples?}
\label{tab:confusion-matrix}
\end{table}

Table \ref{tab:confusion-matrix} shows that \ours is able to correct several previously wrong examples. When \ours corrected previously failing cases, the \nodemoe relied more on mediators, as the average mediator probabilities go up from 0.09 to 0.13 averaged over the datasets. \ours still fails, and more concerning are the cases when previously successful examples now fail. To study this, we annotate 50 random dev samples over the three datasets~(26/24 examples for weakens/strengthens label). For each sample, a human-annotated if the graph had errors. We observe the following error categories:\footnote{Concrete examples in Appendix~\secref{sec:erroranalysisexamples}}
\squishlist
\item \textbf{All nodes off-topic (4\%)}: The graph nodes were not on topic. This (rarely) happens when \ours cannot distinguish the sense of a word in the input question. For instance, \upd = there is a water fountain in the center -- \ours generated based on an incorrect word sense of natural water spring. 
\item \textbf{Repeated nodes (20\%)}: These may be exact or near-exact matches. Node pairs with similar effects tend to be repeated in some samples. 
E.g., the S- node is often repeated with contextualizer C- perhaps because these nodes indirectly affect graph nodes in a similar way. 
\item \textbf{Mediators are uninformative (34\%)}: The mediating nodes are not correct or informative. One source of these errors is when the \hypo and \upd are nearly connected by a single hop, e.g., \hypo = personX pees, and \upd = personX drank a lot of water previously.
\item \textbf{Good graphs are ineffective (42\%)}: These graphs contained the information required to answer the question, but the gating MOE mostly ignored this graph. This could be attributed in part to the observation in the histogram in Figure \ref{fig:graphquesmoedist}, that samples with \textit{weakens} label disproportionately ignore the graph.
\squishend

In accordance with the findings of \citet{rudinger-etal-2020-thinking}, the maximum percentage of errors was in \atomic, in part due to low question quality.


\section{Explainability}
In this section, we analyze the \textit{explainability} of \ours model. 
\citet{jacovi2020towards} note that an explanation should aim towards two complementary goals:
i) Plausibility: provide an interpretation of system outputs that is convincing for humans, and ii) Faithfulness: capture the actual reasoning process of a model.
We discuss how our approach takes a step towards addressing these goals. 

\paragraph{Plausibility}
In a prior work,
\citet{defeasible-human-helps-paper} show that human annotators selectively \textit{picked and chose} parts of the graph that explained a model decision and enabled them in improving on the task of defeasible reasoning.
We show in~\secref{sec:detailedmoeanalysis}, the \moe gate values gives insights into the part of the graph~(contextualizer, mediator, situation node) that the model leveraged to answer a query. Our model thus produces a reasoning chain that is similar to the explanation that humans understand, providing a step towards building inherently plausible models, while also achieving better performance. 

\paragraph{Measuring faithfulness w.r.t. graphs}
Since faithfulness is a widely debated term, we restrict its definition to measure faithfulness w.r.t to the reasoning graph. 
This can be measured by the correlation between the model performance and graph correctness. 
A high correlation implies that the model uses both the graph and query to generate an answer and thus is faithful to the stated reasoning mechanism~(i.e., graphs used to answer a question).
Our analysis reveals this to be the case: in cases where the model answers incorrectly, ~42\% of the graphs were entirely correct (\secref{sec:analysis}).
In contrast, when the model answers correctly, 82\% of the graphs are correct.
In summary, we hope that \ours serves as a step towards building reasoning models that are both plausible and faithful. 
\section{Related work}

\paragraph{Mental Models} Cognitive science has long promoted mental models - coherent, constructed representations of the world - as
central to understanding, communication, and problem-solving \cite{JohnsonLaird1983MentalM,mental-models,Hilton1996MentalMA}.
Our work draws on these ideas, using inference graphs to represent the machine's ``mental model'' of the problem at hand.
Building the inference graph can be viewed as first asking clarification questions about the context before answering. This is similar to self-talk \cite{selftalk} but directed towards eliciting chains of influence.

\paragraph{Injecting Commonsense Knowledge}

Many prior systems use commonsense knowledge to aid question-answering,
e.g., using sentences retrieved from a corpus \cite{Yang2019EndtoEndOQ,Guu2020REALMRL},
or with knowledge generated from a separate source \cite{selftalk,Bosselut2019COMETCT};
and injected either as extra sentences fed directly to the model \cite{Clark2020FromT},
via the loss function \cite{Tandon2018ReasoningAA}, or via attention \cite{Ma2019TowardsGN}.
Unlike prior work, we use conditional language generation techniques to create graphs that are relevant to answering a question.

\paragraph{Encoding Graph Representations}
Several existing methods use graphs as an additional input for commonsense reasoning~\cite{sun2018open,lin2019kagnet,lv2020graph,feng2020scalable,bosselut2021dynamic,ma2020knowledge,kapanipathi2020infusing}.
These methods first retrieve a graph relevant to a question using information retrieval techniques and then encode the graph using graph representation techniques like~\gcn~\cite{kipf2016semi} and graph attention~\cite{velivckovic2017graph}.
Different from these works, we use a graph \textit{generated} from the query for answering the commonsense question.
The graphs consumed by these works contain entities grounded in knowledge graphs like ConceptNet~\cite{speer2017conceptnet}, whereas we perform reasoning over event inference graphs where each node describes an event.
Our best model uses a mixture-of-experts (\moe)~\cite{jacobs1991adaptive} model to pool multi-faceted input.
Prior work has shown the effectiveness of using \moe for graph classification~\cite{zhouexplore,hu2021graph}, cross-lingual language learning~\cite{chen2019multi,gu2018universal}, and model ensemble learning~\cite{fedus2021switch,shazeer2017outrageously}.
To the best of our knowledge, we are the first to use \moe for learning and pooling graph representations for \qa task.
\section{Summary and Conclusion} 

Cognitive science suggests that people form ``mental models'' of a situation to
answer questions about it.
Drawing on those ideas, we have presented a simple instantiation in which the situational model is an inference graph.
Different from \gcn-based models popular in graph learning, we use mixture-of-experts to pool graph representations. Our experiments show that  \moe-based pooling can be a strong~(both in terms of performance and explainability) alternative to \gcn for graph-based learning for reasoning tasks. Our method establishes a new state-of-the-art on three defeasible reasoning datasets.
Overall, our method shows that performance can be improved by guiding a system to ``think about'' a question and explicitly model the scenario, rather than answering reflexively. 
\eat{
We present \ours, a system that achieves a new state-of-the-art on three different defeasible reasoning datasets. We find that primary reason for these gains is that \ours models the question scenario with an inference graph, and judiciously encodes the graph to answer the question. This result is significant, because it shows that performance can be improved by guiding a system to ``think about'' a question and explicitly model the scenario, rather than answering reflexively.

}

\section*{Acknowledgments}
We thank the anonymous reviewers for their feedback.
Special thanks to reviewer 2 for their insightful comments on our \moe formulation.
This material is partly based on research sponsored in part by the Air Force Research Laboratory under agreement number FA8750-19-2-0200. 
The U.S. Government is authorized to reproduce and distribute reprints for Governmental purposes notwithstanding any copyright notation thereon. 
The views and conclusions contained herein are those of the authors and should not be interpreted as necessarily representing the official policies or endorsements, either expressed or implied, of the Air Force Research Laboratory or the U.S. Government.

\bibliographystyle{acl_natbib}
\bibliography{emnlp2021}

\newpage
\clearpage
\appendix
\appendix

\section{Training graph corrector}
\label{sec:trainingdatacorrector}

As mentioned in Section~\secref{sec:graphgen}, the graph generator \gnoisy is trained as a seq2seq model from \wiqa with $\texttt{input} = \text{[Premise] } \V{T}_i \mid \text{[Situation] } \V{S}_i \mid \text{[Hypothesis] } \V{H}_i$, and $\texttt{output} = \V{G}_i$.
Graphs in \wiqa additionally capture the influence that the situation has on the hypothesis.
Denoting this influence label by $y_i$ can be either \textit{helps} or \textit{hurts}

From our experiments, we observe that appending $y_i$ to the training data (from $\texttt{input} = \text{[Premise] } \V{T}_i \mid \text{[Situation] } \V{S}_i \mid \text{[Hypothesis] } \V{H}_i$ to $\texttt{input} = \text{[Premise] } \V{T}_i \mid \text{[Situation] } \V{S}_i \mid \text{[Hypothesis] } \V{H}_i \mid y_i$) reduces repetitions by 13\%.

\begin{figure}[!ht]
\centering
{\includegraphics[width=0.48\textwidth,height=0.10\textheight]{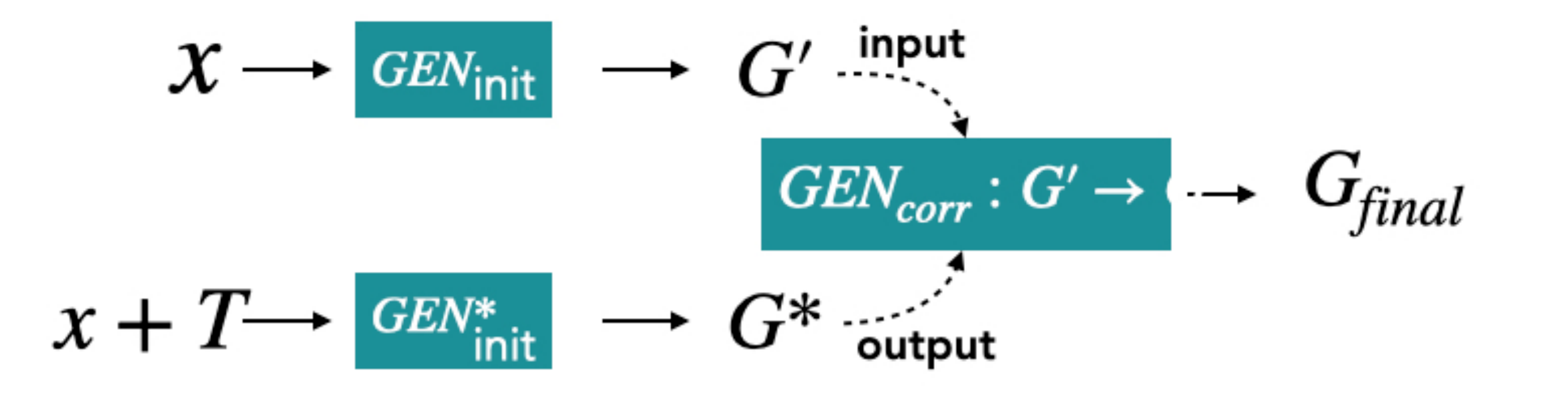}}
\caption{Training data generation to train \gcorr.}
\label{fig:datagen}
\end{figure}

\begin{figure*}[!htb]
\minipage{0.3\textwidth}
  \includegraphics[width=\linewidth]{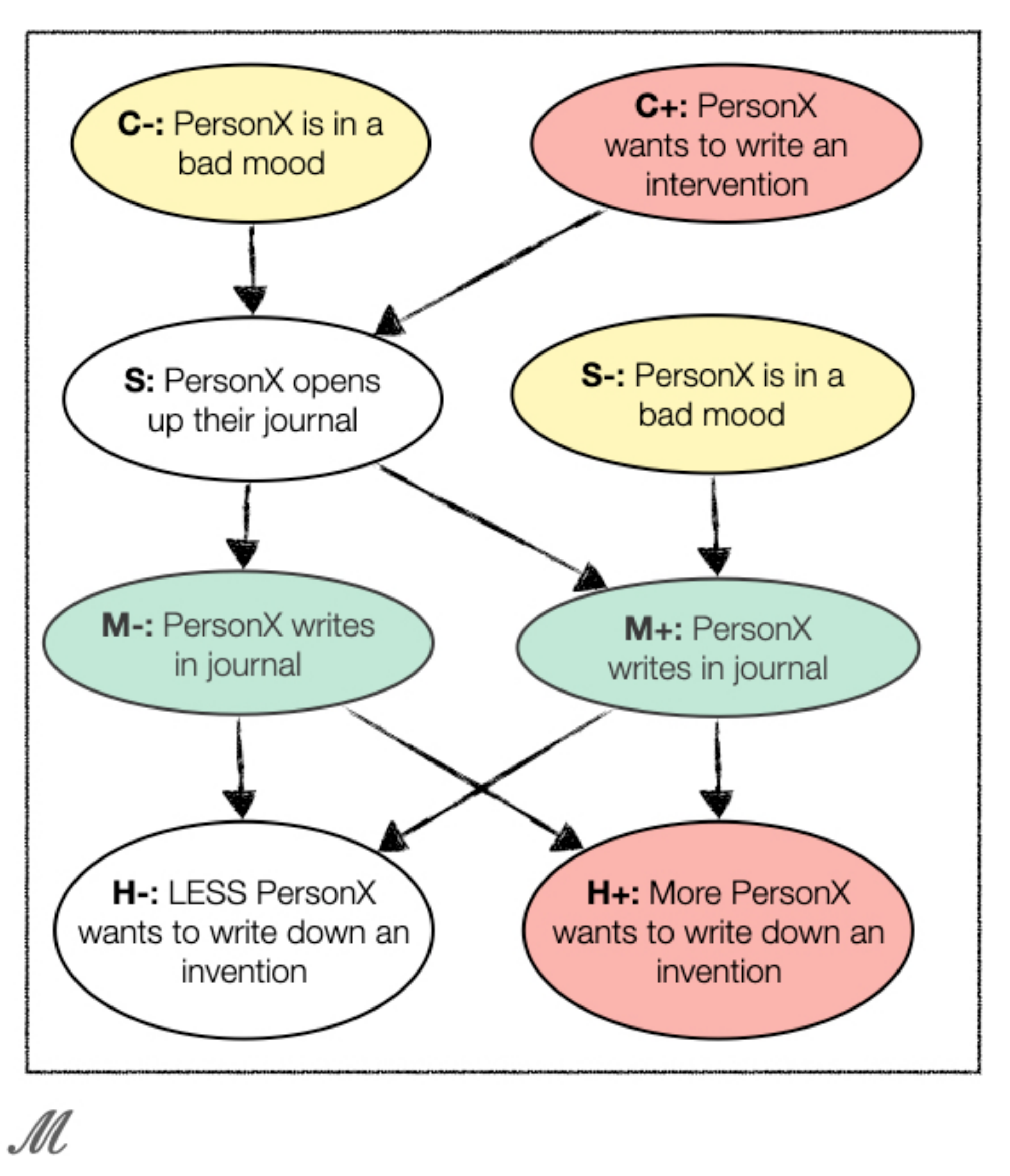}
\endminipage\hfill
\minipage{0.3\textwidth}
  \includegraphics[width=\linewidth]{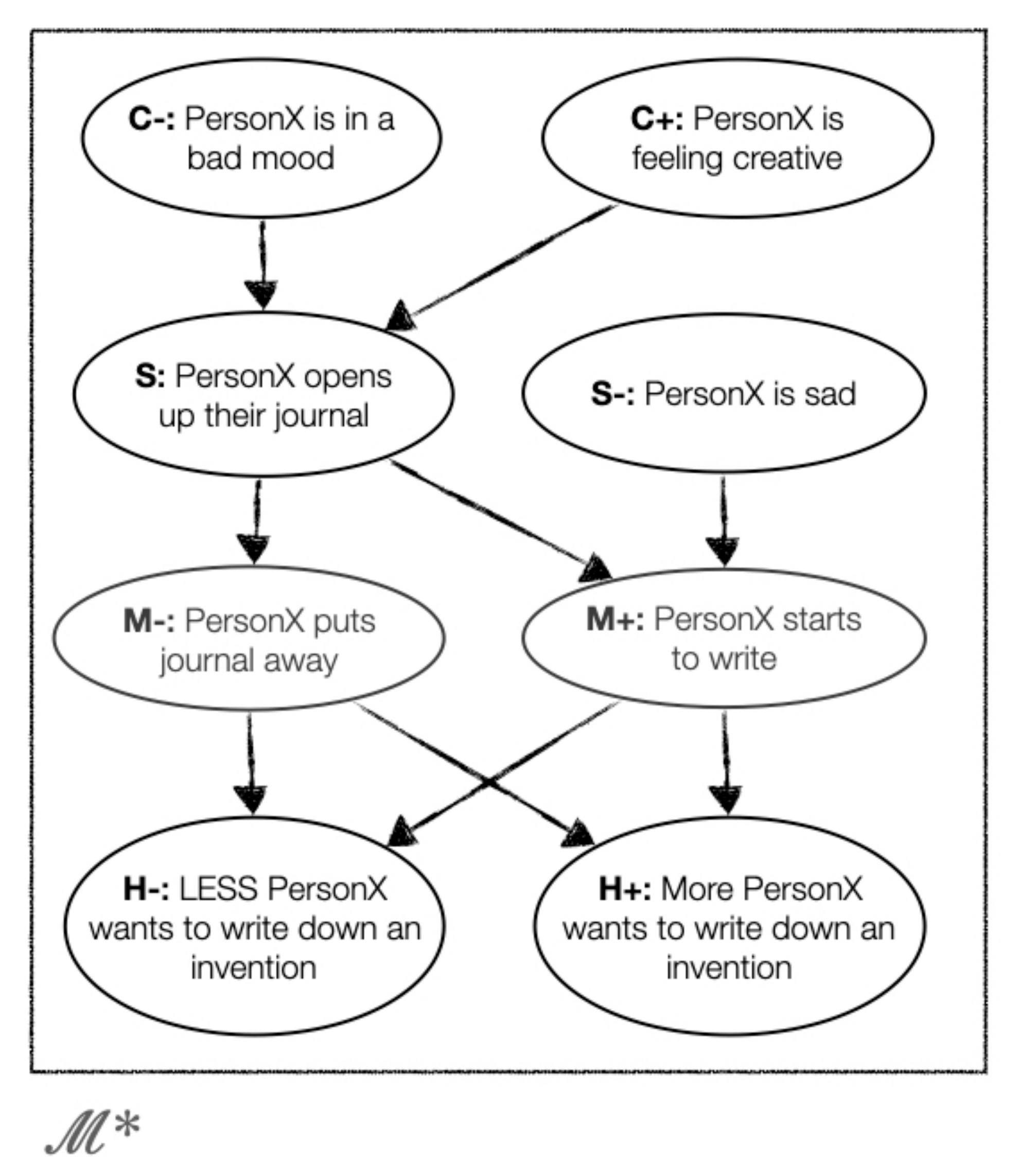}
\endminipage\hfill
\minipage{0.3\textwidth}%
  \includegraphics[width=\linewidth]{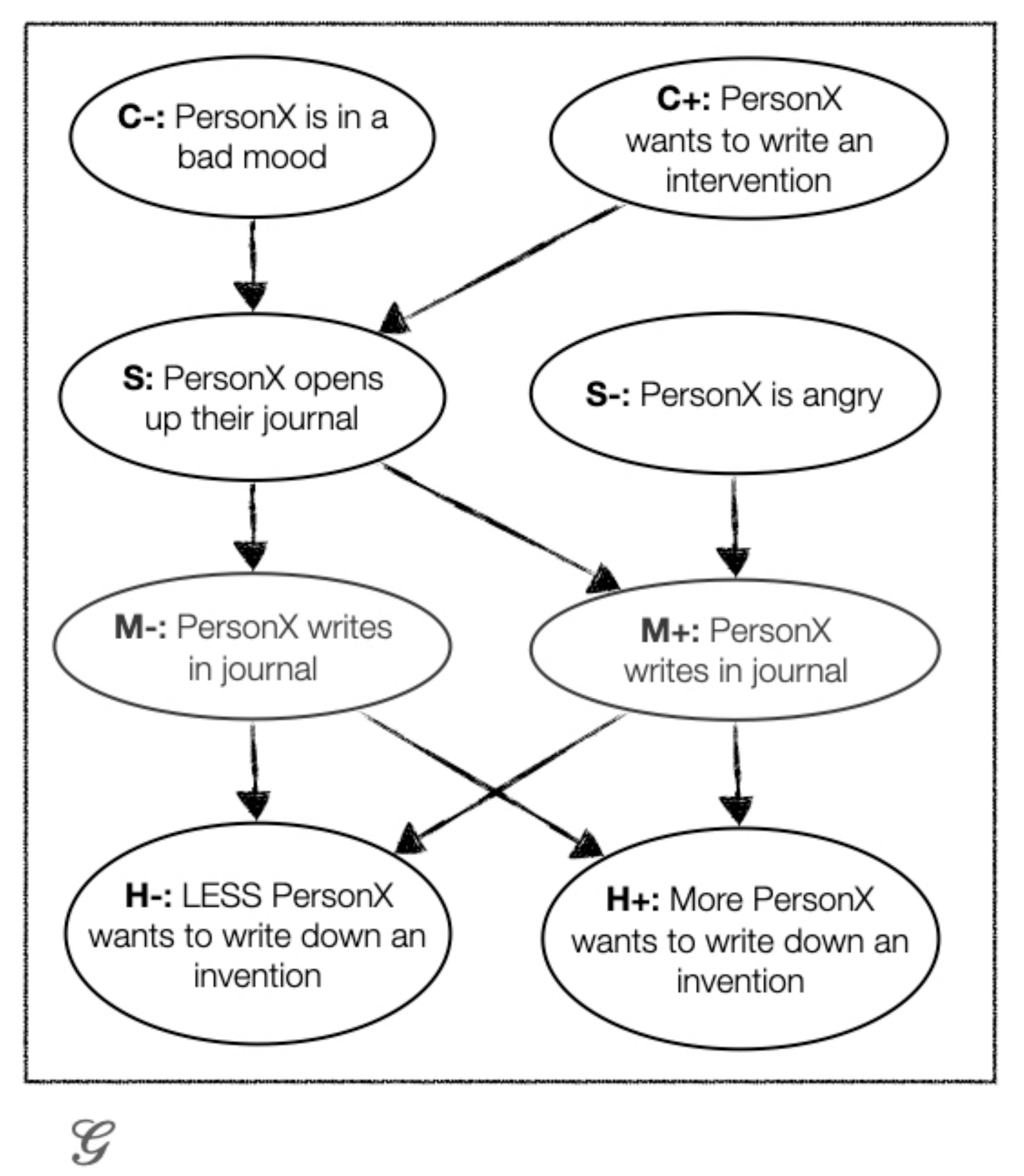}
\endminipage
\caption{The graphs generated by \geninit (left), \genwitht (middle), and \gcorr (right).The input graph has repetitions for nodes $\{C{-}, S{-}\}$,  $\{C{+}, H{+}\}$, and $\{M{-}, M{+}\}$. The corrected graph replaces the repetitions with meaningful labels.}
\label{fig:defeasible_incorrect}
\end{figure*}

We refer to this data generator as \gnoisyt, and the graphs produced by it as \betterg. However, we do not have access to $y$ during test time, and thus \gnoisyt cannot be used directly to produce \betterg for defeasible queries.
We circumvent this by using \gnoisyt to train a graph-to-graph generation model, that takes as input \badg and generates \betterg as output (\badtobetter). We call this system \gcorr.
We give an overview of the process in Figure~\ref{fig:datagen}.
In Figure~\ref{fig:defeasible_incorrect}, we give examples of an intial graph produced by \geninit, the corresponding graph produced by \genwitht, and the graph produced by \gcorr.

\section{\moe gradient analysis}
\label{sec:moegradientanalysis}
\begin{figure*}[!ht]
\centering
{\includegraphics[width=0.78\textwidth,height=0.25\textheight]{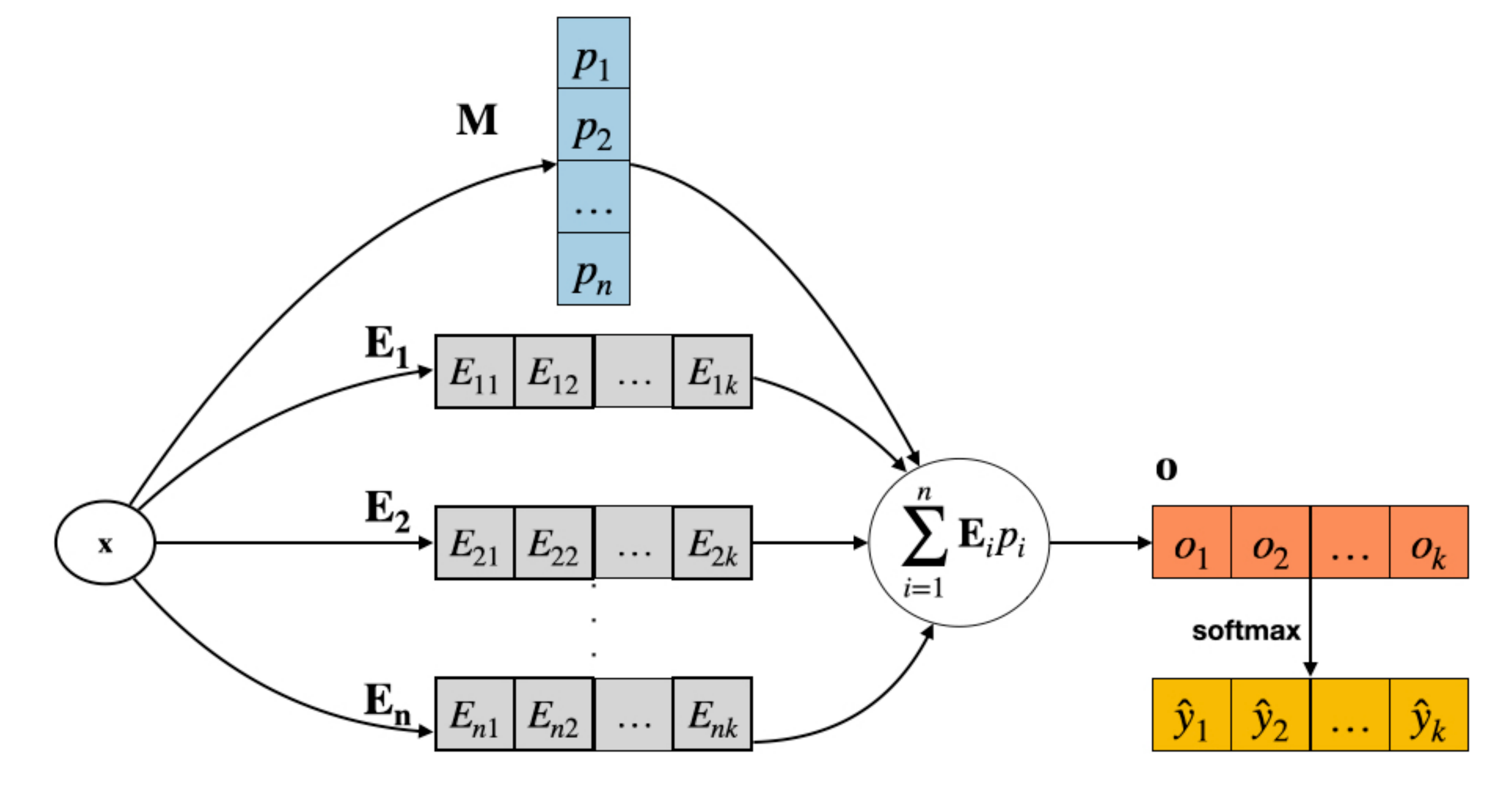}}
\caption{\moe gradient analysis setup: we consider a simple setting where the weighted output of the experts~(using the expert weights $p$) is directly fed to a softmax and is used for generating class probabilities $\hat{y}$.}
\label{fig:moeanalysis}
\end{figure*}

We restate Equation~\ref{eqn:moe} for quick reference:
\begin{align*}
    \V{p} &= \V{M}(\V{x}) \nonumber \\
     \V{o} &= \sum_{i=1}^{n} p_i \V{E_i(x)}
\end{align*}
where we have changed the notation slightly to use $\V{o}$ as the \moe output instead of $\V{y}$.
We also refer to $\V{E_i}(x)$ as $\V{E_i}$.
Further, $o_j = \sum_{i=1}^{n} p_i E_{ij}$.
We present the analysis for a generic multi-class classification setting with $k$ classes, with training done using a cross-entropy loss $\mathcal{L}$~(Figure~\ref{fig:moeanalysis}) Let $\hat{y}_c$ be the normalized probability of the correct class $c$ calculated using softmax: 

\begin{align*}
  \hat{y}_c &= \frac{\exp(o_c)}{\sum_{j=1}^{k}\exp(o_j)}  \\
  &= \frac{\exp(\sum_{i=1}^{n}p_i E_{ic})}{\sum_{j=1}^{k}\exp(\sum_{i=1}^{n}p_i E_{ij})}
\end{align*}


Let $\mathcal{L}$ be the cross-entropy loss:

\begin{align*}
    \mathcal{L} &= -\log \hat{y}_c =  -o_c + \log {\sum_{j=1}^{k}\exp(o_j)} \\
     &= -\sum_{i=1}^{n} p_i E_{ic} + \log {\sum_{j=1}^{k} \exp(\sum_{i=1}^{n} p_i E_{ij})}
\end{align*}

\paragraph{Evaluating $\frac{\partial \mathcal{L}}{\partial p_m}$}
 The derivatives w.r.t. the $m^{th}$ expert gate probability $p_m$ is given by:
\begin{align}
    \frac{\partial \mathcal{L}}{\partial p_m} &= -E_{mc} +  \frac{\sum_{j=1}^{k}E_{mj}\exp(\sum_{i=1}^{n} p_i E_{ij})}{\sum_{j=1}^{k}\exp(\sum_{i=1}^{n} p_i E_{ij})}  \nonumber  \\
    &= -E_{mc} + \sum_{j=1}^{k}\hat{y}_jE_{mj} \nonumber \\
    &= -E_{mc}(1 - \hat{y}_c) + \sum_{j=1,j \not = c}^{k}\hat{y}_j E_{mj}
    \label{eqn:moedbypi}
\end{align}

\paragraph{Evaluating $\frac{\partial \mathcal{L}}{\partial E_{mc}}$}
the derivatives w.r.t. the logits $E_{mc}$ (logit for the correct class by $m^{th}$ expert) is given by:

\begin{align}
    \frac{\partial \mathcal{L}}{\partial E_{mc}} &= -p_m +  \frac{\exp(o_c)p_m}{\sum_{j=1}^{k}\exp o_j}  \nonumber  \\
    &= -p_m(1 - \hat{y}_c)
    \label{eqn:moedbyec}
\end{align}
Equations~\ref{eqn:moedbypi} and \ref{eqn:moedbyec} have natural interpretations: the gradient on both the mixture probability $p_m$ and the logits $E_{mc}$ will be 0~(note that for Equation~\ref{eqn:moedbypi}, $\V{y}^c = 1 \implies \V{y}^j = 0$ for $j \not = c$) when the network makes perfect predictions~($\hat{y}_c = 1$).
As noted by~\citet{jacobs1991adaptive}~(Section 1), this might cause the network to specialize slower, as the gradient will be small for experts that are helping in making the correct prediction. 
They suggest a different loss function that promotes faster specialization by redefining the error function in terms of a mixture distribution, with the mixture weights supplied by the $p_i$ terms.
Analyzing the effect of loss function for applications where the \moe is used to pool representations remains an interesting future work.

\section{Hyperparameters}
\label{sec:hyperparams}
\paragraph{Training details}
All of our experiments were done on a single Nvidia GeForceRTX 2080 Ti.
We base our implementation on PyTorch~\cite{paszke2017automatic} and also use PyTorch Lightning~\cite{falcon2019pytorch} and Huggingface~\cite{wolf2019huggingface}.
The gates and the experts in our \moe model were a single layer MLP. For the experts, we set the input size set to be the same as output size.
Table~\ref{tab:hyperparams} shows the parameters shared by all the methods, and \ref{tab:gcn-hyperparams} shows the hyperparameters applicable to \gcn encoder.

\begin{table}
\centering
\small
\begin{tabular}{lr} \\ \toprule
Hyperparameter                & Value    \\ \midrule
Pre-trained model &  \roberta -base \\
Learning rate                 & 2e-5     \\ 
Gradient accumulation batches & 2        \\
Num epochs                    & 30       \\
Optimizer & AdamW \\
Dropout & 0.1 \\
Learning rate scheduling      & linear   \\
Warmup                        & 3 epochs \\
Batch size                    & 16       \\
Weight decay                  & 0.01     \\
Gradient clipping             & 1.0     \\  \bottomrule
\end{tabular}
\caption{General hyperparameters used by all the models.}
\label{tab:hyperparams}
\end{table}

\begin{table}
\centering
\begin{tabular}{lr} \\  \toprule
Hyperparameter            & Value \\ \midrule
\# Layers                 & 2     \\
Layer dropout             & 0.1   \\
Number of attention heads & 1     \\
Attention dimension       & 256  \\  \bottomrule
\end{tabular}
\caption{Hyperparameters specific to \gcn.}
\label{tab:gcn-hyperparams}
\end{table}



\section{Schema of an influence graph}
\label{sec:igraphschema}
\begin{figure}[!ht]
\centering
{\includegraphics[scale=0.25]{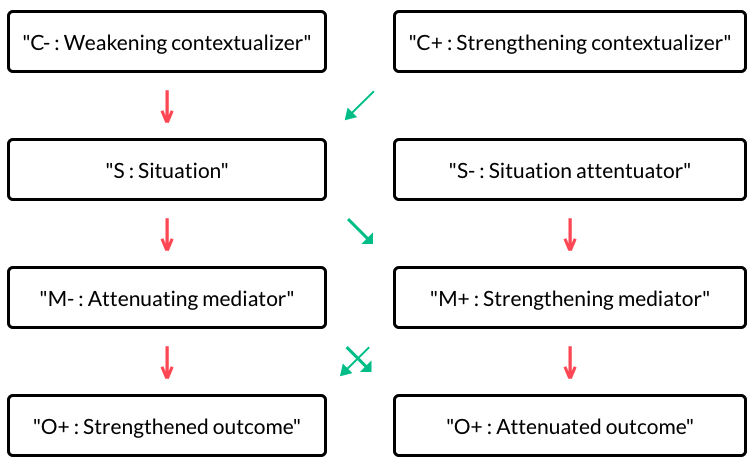}}
\caption{Schema of an inference graph.}
\label{fig:inference-graph-schema}
\end{figure}

Figure~\ref{fig:inference-graph-schema} shows the skeleton of an influence graph.

\section{Runtime Analysis}
\label{sec:runtime}
Finally, we discuss the cost-performance tradeoffs for various encoding mechanisms~(Table~\ref{tab:tradeoffs}).
As Table~\ref{tab:tradeoffs} shows, both \gcn and \moe take about 7\% more number of parameters than the \str encoding scheme and have about 2x the runtime.
Further, as we use one expert per node, the number of parameters scales linearly with the number of nodes. 
While this is not prohibitive in our setting (each graph has a small number of nodes), our analysis shows that the behavior of the nodes that have similar semantics is correlated, indicating that the experts for those nodes can share parameters. 
Alternatively, \moe with more than two layers~\cite{jordan1995convergence} can also help in scaling the number of parameters only logarithmically with the number of nodes.

\begin{table}[!ht]
\centering
\begin{tabular}{llll} \toprule
Method & \str & \gcn & \moe   \\ \midrule
\#Params & 124M & 131M & 133M \\
Runtime & 0.17 &  0.47 & 0.40  \\ \bottomrule
\end{tabular}
\caption{Number of parameters in the different encoding methods. Runtime reports the number of seconds to process one training example.}
\label{tab:tradeoffs}
\end{table}




\section{Error Analysis Examples}
\label{sec:erroranalysisexamples}
\begin{figure*}[!ht]
\centering
{\includegraphics[scale=0.45]{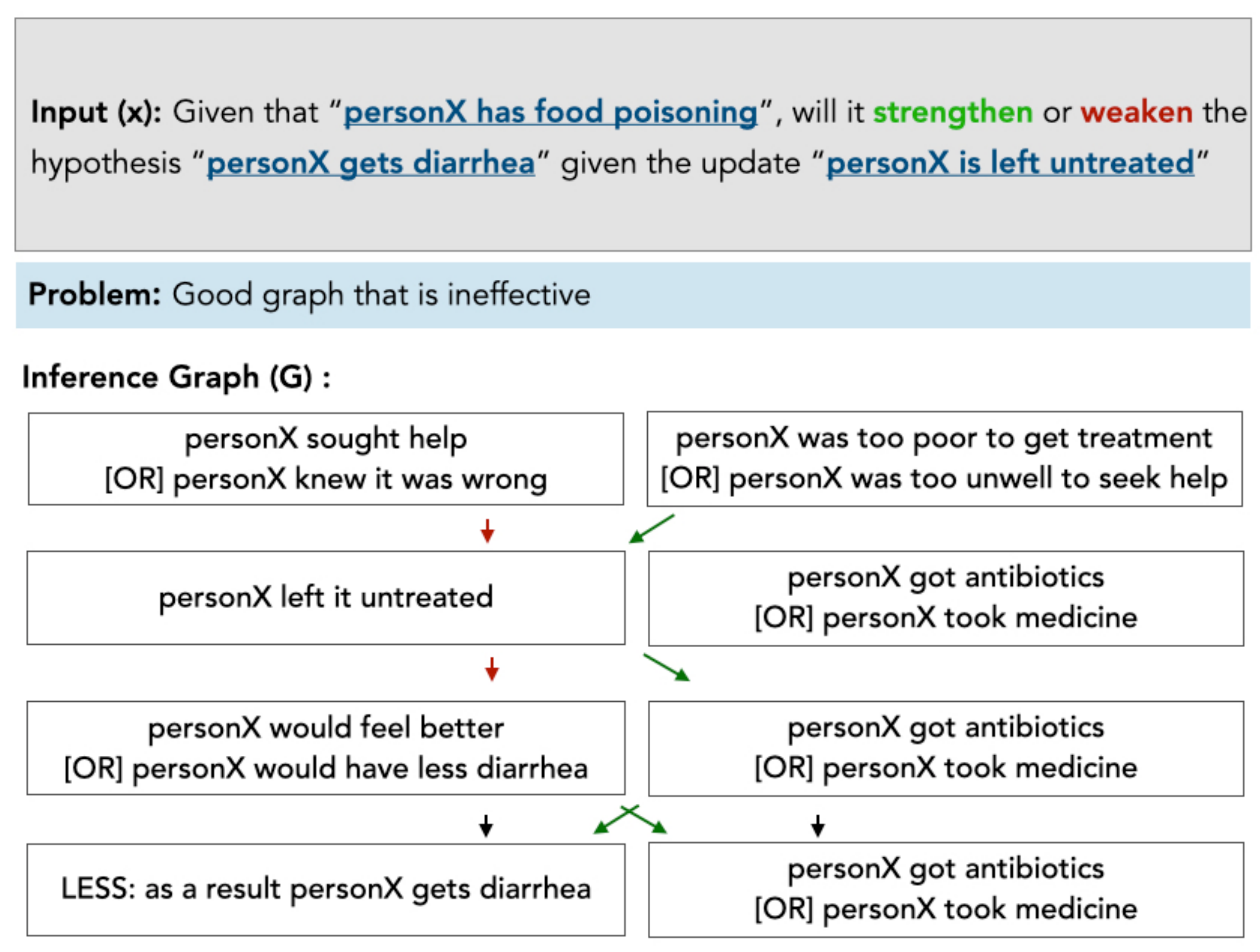}}
\caption{Example of a failure case: A good graph is unused. Example from \atomic dev set.}
\label{fig:ig-atomic-good-graph-unused}
\end{figure*}

\begin{figure*}[!ht]
\centering
{\includegraphics[scale=0.45]{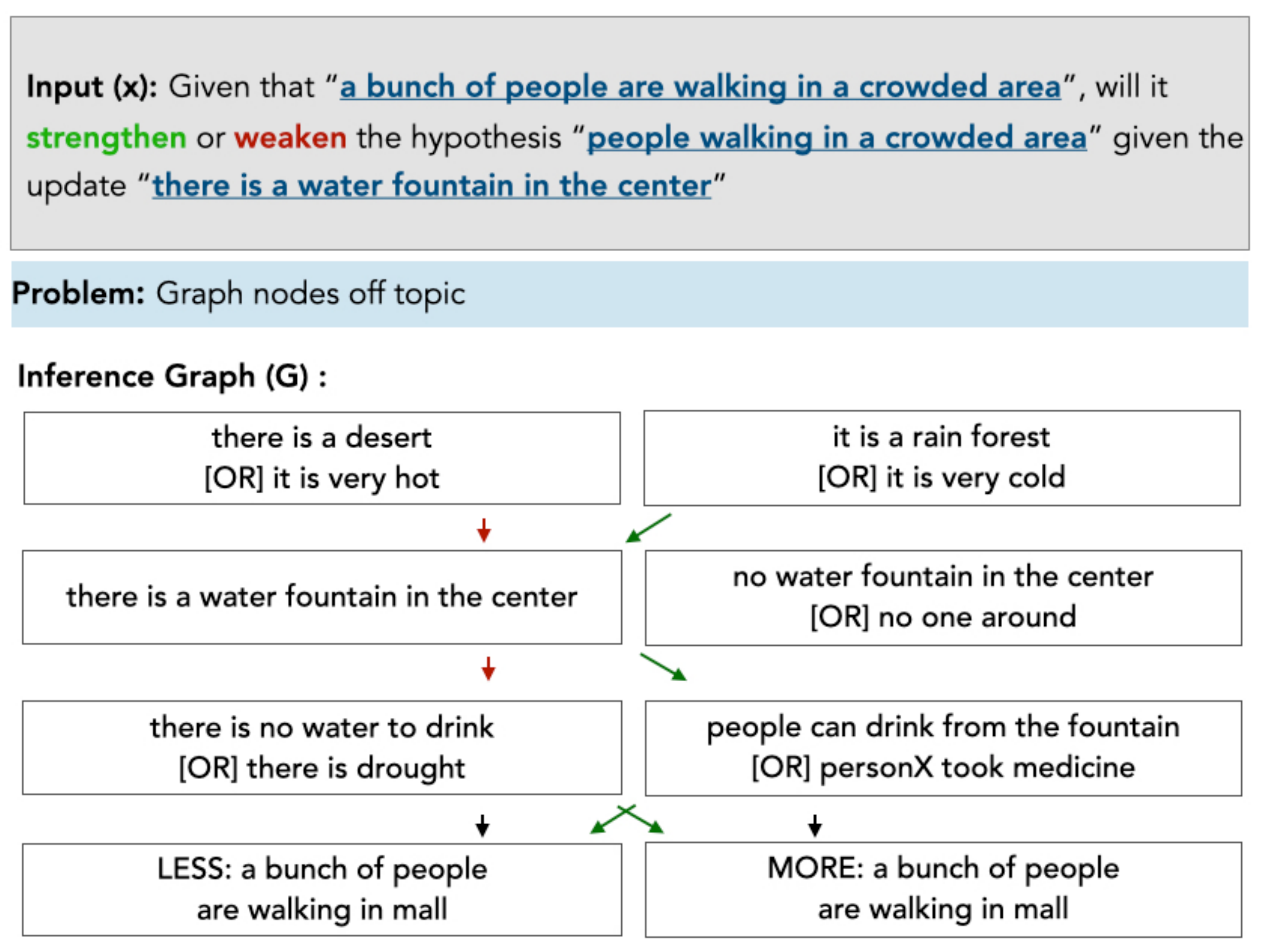}}
\caption{Example of a failure case: The generated graph is off topic (wrong sense of water fountain is used). Example from \snli dev set.}
\label{fig:ig-snli-off-topic}
\end{figure*}

\begin{figure*}[!ht]
\centering
{\includegraphics[scale=0.45]{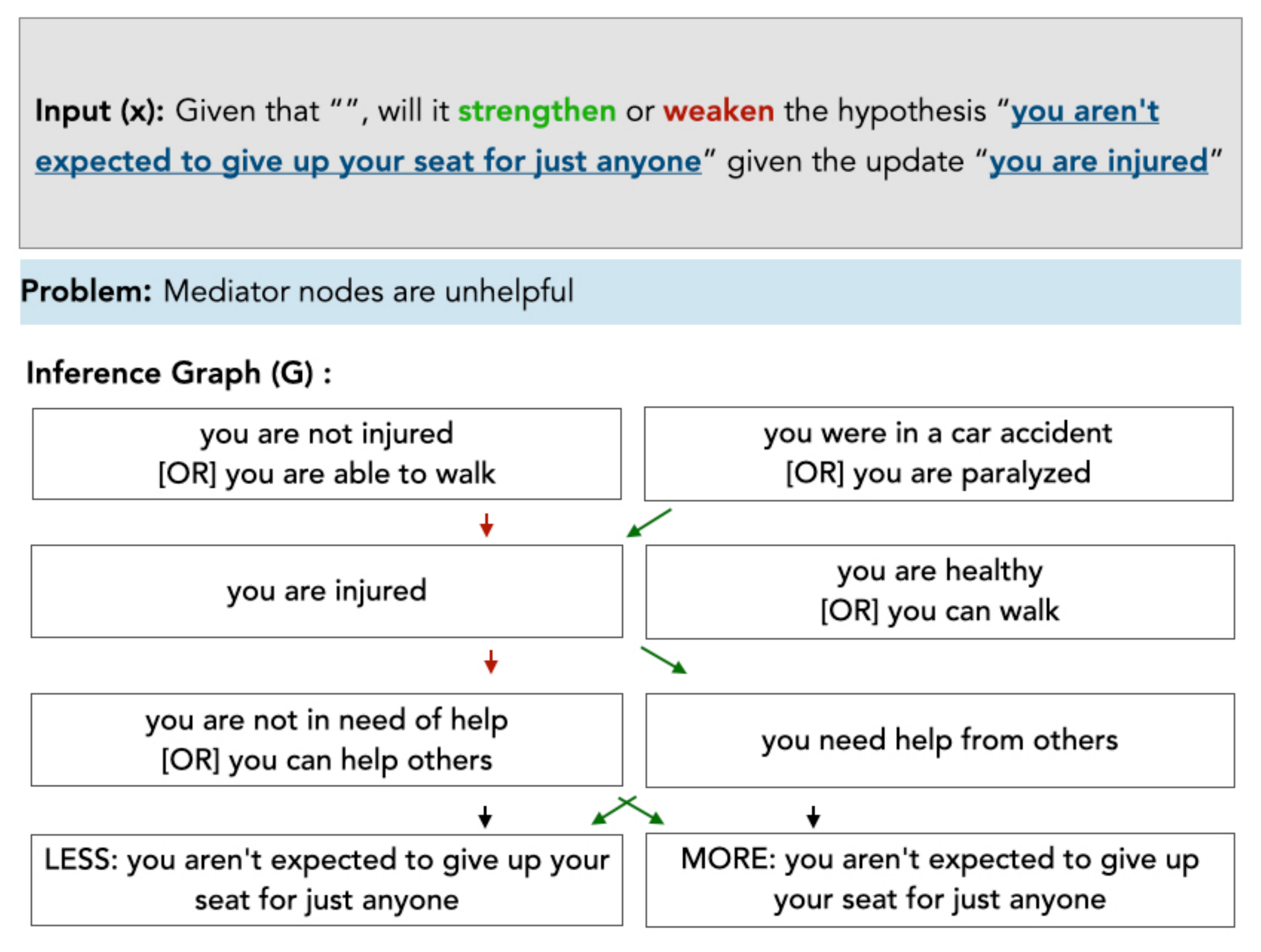}}
\caption{Example of a failure case: The mediator nodes (second last level in the graph) are unhelpful. Example from \social dev set.}
\label{fig:ig-social-mediator-wrong}
\end{figure*}

We show three examples with different types of errors. These examples are taken from Dev set, and these are for the cases where \ours introduced a wrong answer, while baseline answered this correctly without the graph.
\squishlist
\item Figure \ref{fig:ig-atomic-good-graph-unused} shows a failure case when a good graph is unused. Example from \atomic dev set.
\item Figure \ref{fig:ig-snli-off-topic} shows a failure case when an off topic graph is produced due to confusion in the sense of water fountain. Example from \snli dev set.
\item Figure \ref{fig:ig-social-mediator-wrong} shows a failure case when the mediator is wrong. Example from \social dev set.
\squishend

\section{Significance Tests}
\label{sec:statsig}
We perform two statistical tests for verifying our results: i) The micro-sign test (s-test)~\cite{yang1999re}, and ii) McNermar's test~\cite{dror2018hitchhiker}.

\begin{table}[ht]
\centering
\begin{tabular}{lrr} \toprule
Dataset & s-test & McNemar's test \\
\midrule
\atomic & 5.07e-05 & 1.1e-04 \\
\snli & 2.65e-05 & 6.5e-05 \\ 
\social & 1.4e-04 & 3.2e-04 \\ \bottomrule
\end{tabular}
\caption{p-values for the three datasets and two different statistical tests while comparing the results with and without graphs (Table~\ref{tab:overall-main-result}). As the p-values show, the results in Table~\ref{tab:overall-main-result} are highly significant}
\label{tab:statanalysis}
\end{table}

\begin{table}[ht]
\centering
\begin{tabular}{lrr} \toprule
Dataset & s-test & McNemar's test \\
\midrule
\atomic & 0.001 & 0.003676 \\
\snli & 0.01 & 0.026556 \\ 
\social & 0.06 & 0.146536 \\ \bottomrule
\end{tabular}
\caption{p-values for the three datasets and two different statistical tests while comparing the results with noisy vs. cleaned graphs (Table~\ref{tab:better_graphs_better_perf}).}
\label{tab:statanalysisggen2}
\end{table}

\begin{table}[!ht]
\centering
\begin{tabular}{lrrr}
\toprule
 & \atomic & \snli & \social \\
\midrule
\str & 0.13  & 1.8e-06 & 8.7e-06  \\
\gcn & 0.006 & 1.31e-05 & 0.03 \\
\bottomrule
\end{tabular}
\caption{p-values for the s-test for Table~\ref{tab:better_encoding_better_perf_1}.}
\label{tab:stat_analysis_better_encoding_better_perf_1}
\end{table}

\begin{table}[!ht]
\centering
\begin{tabular}{lrrr}
\toprule
 & \atomic & \snli & \social \\
\midrule
\str & 0.28 & 4e-06 & 2e-05 \\
\gcn & 0.015127 & 3.2e-05 & 0.06 \\
\bottomrule
\end{tabular}
\caption{p-values for the McNemar's for Table~\ref{tab:better_encoding_better_perf_1}.}
\label{tab:stat_analysis_better_encoding_better_perf_2}
\end{table}

\newpage
\clearpage

\section{Description of \gcn encoder}
\label{sec:gcn_pooling}

\begin{figure*}[!ht]
\centering
{\includegraphics[width=0.80\textwidth,height=0.24\textheight]{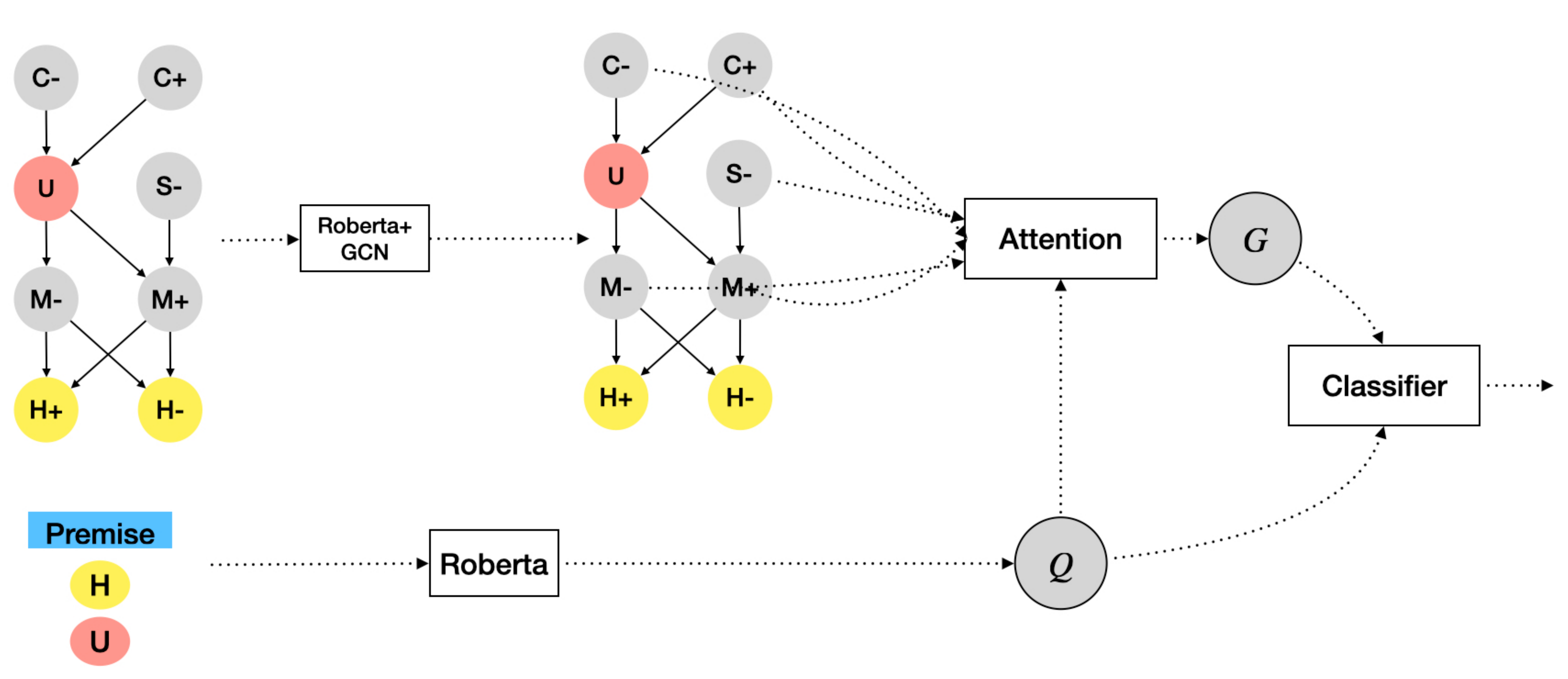}}
\caption{Overview of the \gcn encoder.}
\label{fig:gcn}
\end{figure*}

We now describe our adaptation of the method by~\citet{lv2020graph} to pool $\nodeset$ into \hg using \gcn.
Figure~\ref{fig:gcn} captures the overall design.
\paragraph{Refining node representations}
The representation for each node $v \in \V{V}$ is first initialized using:
$$\V{h}_\V{v}^0 = \V{W}^{0}\V{h}_\V{v}$$

Where $\V{h}_\V{v} \in \real{d}$ is the node representation returned by the \lm, and  $\V{W}^{0} \rdim{d \times k}$.
This initial representation is then refined by running L-layers of a GCN~\cite{kipf2016semi}, where each layer $l + 1$ is updated by using representations from the $l^{th}$ layer as follows:

\begin{align}
\V{h}_v^{(l + 1)} &= \sigma\left(\frac{1}{|\V{A}(v)|}{\sum_{w \in \V{A}(v)}\V{W}^{l}\V{h}_w^l}  + \V{W}^{l}\V{h}_v^l\right) \nonumber \\
\V{H}^{L} &= [\V{h}_0^L; \V{h}_1^L; \ldots; \V{h}_{|\V{V}| - 1}^L] \\
\label{eqn:gcnn}
\end{align}

Where $\sigma$ is a non-linear activation function, $\V{W}^{l} \rdim{k \times k}$ is the GCN weight matrix for the $l^{th}$ layer, $\V{A}(v)$ is the list of neighbors of a vertex $v$, and $\V{H}^{L} \rdim{|\V{V}| \times k}$ is a matrix of the $L^{th}$ layer representations the $|\V{V}|$ nodes such that $\V{H}^{L}_i = \V{h}_i^L$.

\paragraph{Learning graph representation}
We use multi-headed attention~\cite{vaswani2017attention} to combine the query representation $\V{h}_\V{Q}$ and the nodes representations $\V{H}^{L}$ to learn a graph representation $\V{h}_\V{G}$.
The multiheaded attention operation is defined as follows:
\begin{align}
\V{a}_i &=\text{softmax}\left(\frac{(\V{W^q_i}\V{h}_\V{Q})(\V{W^k_i}\V{H}^{L})^{T}}{\sqrt{d}}\right) \nonumber \\
\text{head}_i &=\V{a}_i (\V{W^v_i}\V{H}^{L}) \nonumber \\
\V{h}_\V{G} &=  Concat(head_1, \ldots, head_h)\V{W}^O \nonumber \\
& =  \text{MultiHead}(\V{h}_Q, \V{H}^{L})
\label{eqn-attn}
\end{align}

Where $h$ is the number of attention heads, $\V{W}^q_i, \V{W}^k_i, \V{W}^v_i \rdim {k \times d}$ and $\V{W}^O \rdim{hd \times d}$.

Finally, the graph representation generated by the the MultiHead attention $\V{h}_\V{G} \rdim {n}$ is concatenated with  with the question representation $\V{h}_\V{Q}$ to get the prediction:

$$\hat{y} = \text{softmax}([\V{h}_\V{G}, \V{h}_\V{Q}]\V{W}_{out})$$
where $\V{W}_{out} \rdim{d \times 2}$ is a single linear layer MLP.
\section{All results}

Our experiments span two types of graphs  (\badg, \corrg), three datasets (\snli, \social, \atomic), and three graph encoding schemes (\str, \gcn, \moe). Table~\ref{tab:allresults} above shows the results on all 18 combinations of $\{\text{graph types}\} \times \{\text{datasets}\} \times \{\text{graph encoding schemes}\}$

\begin{table}[!ht]
\centering
\begin{tabular}{p{1.5cm}p{1.2cm}p{1cm}p{1.2cm}}
\toprule
Dataset                  & Encoder & Graph Type  & Accuracy \\ \midrule
\multirow{7}{*}{\atomic} &       & n/a    &          \\
                         & \str  & \badg  & 78.78    \\
                         & \str  & \corrg & 79.48    \\
                         & \gcn  & \badg  & 78.25    \\
                         & \gcn  & \corrg & 78.85    \\
                         & \moe  & \badg  & 78.83    \\
                         & \moe  & \corrg & \textbf{80.15 }   \\ \midrule
\multirow{7}{*}{\snli}   &       & n/a    &          \\
                         & \str  & \badg  & 82.16    \\
                         & \str  & \corrg & 83.11    \\
                         & \gcn  & \badg  & 82.63    \\
                         & \gcn  & \corrg & 83.09    \\
                         & \moe  & \badg  & 83.83    \\
                         & \moe  & \corrg & \textbf{85.59}    \\ \midrule
\multirow{7}{*}{\social} &       & n/a    & 87.6     \\
                         & \str  & \badg  & 86.75    \\
                         & \str  & \corrg & 87.24    \\
                         & \gcn  & \badg  & 87.92    \\
                         & \gcn  & \corrg & 88.12    \\
                         & \moe  & \badg  & 88.45    \\
                         & \moe  & \corrg & \textbf{88.62}    \\ \bottomrule
\end{tabular}
\caption{Results for different combinations of graph encoder, graph type.}
\label{tab:allresults}
\end{table}

\section{Graph-augmented defeasible reasoning algorithm}

In Algorithm~\ref{alg:alg-1}, we outline our graph-augmented defeasible learning process.

\SetKwInput{KwGiven}{Given}
\SetKwInput{KwInit}{Init}
\begin{algorithm}
\SetAlgoLined
\KwGiven{A language model \lm, defeasible query with graph $(\V{x}, \V{G})$.}
\KwResult{Result for the query.}
\tcp{Encode query}
$\V{h}_\V{Q} \gets \mathcal{L}(\V{x})$\;
\tcp{encode nodes of \gengraph}
$\V{h}_\V{V}\gets \mathcal{L}(\V{v} \in \V{G})$ \;
\tcp{\texttt{MOE1}: Combine nodes}
$ \V{h}_{\V{G}}\gets $ Equation~\ref{eqn:moegraph}\;
\tcp{\texttt{MOE2}: Combine $Q$, $G$ }
$ \V{h}_{\V{y}}\gets $ Equation~\ref{eqn:moenodequery}\;
\KwRet{\texttt{softmax}(\texttt{MLP}($\V{h}_{\V{y}}$))}
\caption{Graph-augmented defeasible reasoning using \moe.}
\label{alg:alg-1}
\end{algorithm}

\end{document}